\PassOptionsToPackage{dvipsnames}{xcolor}
\documentclass[lettersize,journal]{IEEEtran}
\usepackage{amsmath,amsfonts}
\usepackage{algorithmic}
\usepackage{algorithm}
\usepackage{array}
\usepackage[caption=false,font=normalsize,labelfont=sf,textfont=sf]{subfig}
\usepackage{textcomp}
\usepackage{stfloats}
\usepackage{url}
\usepackage{verbatim}
\usepackage{graphicx}
\usepackage{cite}
\usepackage[colorlinks=true, citecolor=blue]{hyperref}
\usepackage{booktabs}
\usepackage{bbding}
\usepackage{xcolor}
\usepackage{wrapfig}
\usepackage{multirow}
\usepackage[table]{xcolor}
\usepackage[caption=false]{subfig}
\usepackage{makecell}
\usepackage{orcidlink}

\begin{document}

\title{LiveStarPro: Proactive Streaming Video Understanding with Hierarchical Memory for Long-Horizon Streams}

\author{Zhenyu Yang\,\orcidlink{0009-0005-5298-0543}, Kairui Zhang\orcidlink{0009-0000-9155-0988}, Bing Wang\orcidlink{0009-0000-1018-6425}, Shengsheng Qian\,\orcidlink{0000-0001-9488-2208},~\IEEEmembership{Member,~IEEE,} and \\ Changsheng Xu\,\orcidlink{0000-0001-8343-9665},~\IEEEmembership{Fellow,~IEEE}


}

\markboth{Preprint, 2026}
{Yang \MakeLowercase{\textit{et al.}}: LiveStarPro: Proactive Streaming Video Understanding with Hierarchical Memory}


\maketitle

\begin{abstract}
Despite the remarkable progress of Video Large Language Models (Video-LLMs), current online architectures still struggle to simultaneously process continuous video streams, decide autonomously when to respond, and preserve long-horizon contextual memory. These obstacles undermine real-time responsiveness and cause severe forgetting throughout prolonged interactions. In this work, we introduce \textbf{LiveStarPro}, a live streaming assistant that is designed for proactive video understanding over long-horizon streams.
The design of LiveStarPro rests on three complementary components. The first component is \textbf{Streaming Verification Decoding (SVeD)}, an inference framework that identifies the appropriate response timing through single-pass perplexity verification, thereby eliminating the dependency on explicit silence tokens. The second component is \textbf{Streaming Causal Attention Masks (SCAM)}, a training strategy that enforces incremental video-language alignment over variable-length streams so that dynamic verification becomes feasible. The third component is \textbf{Tree-Structured Hierarchical Memory (TSHM)}, a recursive memory architecture that organizes evicted historical information into event chains and consequently enables efficient retrieval from effectively unbounded video streams without saturating the active context window.
To facilitate a comprehensive evaluation under realistic online conditions, we further present \textbf{OmniStarPro}, a large-scale benchmark that spans 15 diverse real-world scenarios and that extends to hour-scale streams for the assessment of long-term recall. Extensive experiments demonstrate that LiveStarPro consistently surpasses existing methods, attaining a 28.9\% improvement in semantic correctness and an 18.2\% reduction in timing error relative to prior online Video-LLMs, while its streaming key-value cache further yields a 1.58$\times$ inference speedup over the same model without caching. The model and the code are publicly available at \url{https://github.com/sotayang/LiveStarPro}.
\end{abstract}


\section{Introduction}
\IEEEPARstart{T}{he} swift advancement of Large Vision-Language Models (LVLMs)~\cite{chen2024expanding, Qwen2VL, yao2024minicpm, zhang2024internlm, glm2024chatglm} has substantially transformed the field of Video Large Language Models (Video-LLMs)~\cite{ataallah2024minigpt4, maaz2023video,li2023videochat,yang2023vid2seq,wang2022internvideo}. Recent architectures exhibit notable competence across complex multimodal tasks, owing to sustained progress in spatial-temporal reasoning~\cite{cheng2024videollama, liu2024oryx, ren2024timechat}, memory-augmented processing~\cite{zhang2024flash, song2024moviechat+, he2024ma}, and long-context comprehension~\cite{wang2024longllavascalingmultimodalllms, longvila, zhang2024long}. Nevertheless, these achievements have been obtained predominantly under offline settings, in which models operate over finite and pre-recorded video sequences.

\begin{figure*}[t]
\centering
\includegraphics[width=1.0\textwidth]{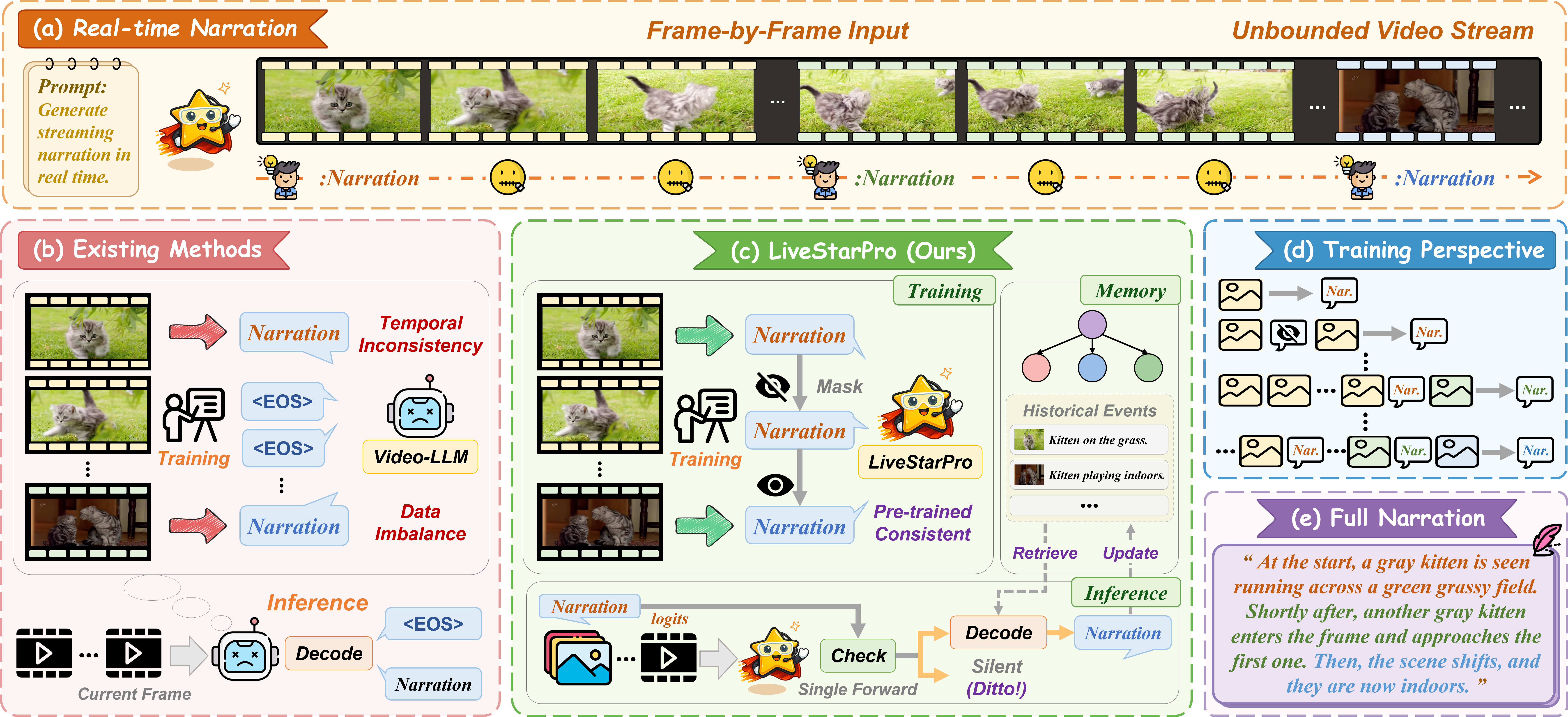} 
\caption{\small \textbf{Illustration of online video understanding.} 
(a) Taking the RNG task as an example, online video understanding requires Video-LLMs to continuously process unbounded video streams and respond only at appropriate moments.
(b) Existing EOS-based methods suffer from data imbalance and temporal inconsistency, leading to unstable training and suboptimal online inference.
(c)-(e) LiveStarPro establishes an effective proactive  response-silence framework through training (SCAM), inference (SVeD), and memory (TSHM), enabling coherent and context-aware real-time narration without compromising basic video understanding capabilities.}
\label{fig:overview}
\end{figure*}

Driven by such progress, growing research attention has been directed toward live streaming assistants for online video understanding~\cite{chen2024videollm, wu2024videollm, li2025lion, ding2025streammind,wang2024videollm,qian2024streaming}. Unlike their offline counterparts, these agents must remain always-on, reason continually as time unfolds, and deliver feedback in real time.
As depicted in Fig.~\ref{fig:overview}(a), a central difficulty in this setting arises from the need to process continuous frame-by-frame inputs while autonomously deciding the most appropriate moment to respond~\cite{qian2024streaming,zhou2024streaming,xiong2025streaming}.
To operate effectively, such models must be endowed with proactive temporal decision-making, so that responses are emitted only at contextually relevant moments rather than as repetitive outputs or default refusals (e.g., ``I do not know'') for every incoming frame.
As an early attempt, VideoLLM-online~\cite{chen2024videollm} proposed a streaming EOS (End-Of-Sequence) prediction mechanism (Fig.~\ref{fig:overview}(b)) that continually consumes the video stream and emits EOS tokens to mark silence intervals conditioned on user queries.
Building upon this idea, subsequent studies such as VideoLLM-MoD~\cite{wu2024videollm} and LION-FS~\cite{li2025lion} refined the EOS-based framework to improve computational efficiency and response accuracy.


\IEEEpubidadjcol
Despite these improvements, the EOS-driven paradigm remains fundamentally limited, giving rise to four structural deficiencies.
(1) \textbf{Severe Data Imbalance}: The frames requiring silence far outnumber those triggering a response~\cite{ding2025streammind}. A 90-second stream sampled at 2 FPS produces 180 frames; if only 6 events deserve a spoken response, the remaining 174 frames map to silence, a response-to-silence ratio of approximately 1:29.
(2) \textbf{Temporal Inconsistency}: As shown in Fig.~\ref{fig:overview}(b), temporally adjacent and visually near-identical frames frequently receive contradictory targets, where one triggers a detailed narration while the next demands an immediate EOS, hindering convergence during fine-tuning.
(3) \textbf{Objective Misalignment}: Standard pre-training optimizes semantic alignment between visual features and textual descriptions, whereas the silence mechanism forces a mapping from rich visual evidence to a null EOS token, conflicting with the meaningful cross-modal correspondence underlying pre-training.
(4) \textbf{Vocabulary Degradation}: Treating the EOS token as an ordinary vocabulary entry lets its high frequency contaminate the semantic space, introducing ambiguity and distorting the natural probability distribution over meaningful tokens.
Together, these deficiencies impede optimization and weaken video-language alignment, ultimately eroding the core video understanding capability of the model.
We therefore contend that silence should not be a learned prediction target, but a derived state verified through the model's confidence. It shifts the central problem from predicting silence to verifying relevance, leading to \textbf{Challenge 1}: \textit{How to establish an effective response-silence framework during both training and inference while preserving video-language alignment?}

Beyond response timing, deploying an always-on assistant exposes fundamental limitations in the memory management of existing online Video-LLMs, which predominantly rely on simple sliding window mechanisms suffering from three inherent obstacles.
(1) \textbf{Context Window Saturation}: Irrespective of the context size (e.g., 8192 tokens), a continuous and potentially unbounded stream eventually exceeds the model capacity, forcing a trade-off between recent observations and long-term history.
(2) \textbf{Catastrophic Forgetting}: Common eviction strategies such as First-In-First-Out (FIFO) discard historical content indiscriminately once the buffer is full, permanently removing access to past events and preventing answers to queries about distant history (e.g., ``What did the person pick up an hour ago?'').
(3) \textbf{Retrieval Inaccessibility}: Existing streaming architectures treat the context buffer as a passive container without an explicit mechanism to recall evicted information. Consequently, historical events outside the active window become inaccessible during inference, preventing the model from relating current observations to relevant long-term context.
These obstacles motivate \textbf{Challenge 2}: \textit{How can long-horizon video streams be processed with efficient long-term memory retrieval while maintaining contextual consistency beyond a fixed context window?}

In parallel, the progress of online Video-LLMs is further constrained by the limited diversity of training data and the narrow scope of existing evaluation benchmarks, which jointly fail to capture the full spectrum of real-world streaming applications. Most representative models~\cite{chen2024videollm,wu2024videollm,li2025lion} rely heavily on first-person videos drawn from Ego4D~\cite{grauman2022ego4d}. Although StreamMind~\cite{ding2025streammind} broadens the domain to sports through SoccerNet~\cite{giancola2018soccernet}, the overall coverage of real-world contexts remains sparse. A comparable limitation is observed in current evaluation protocols. Recent benchmarks such as SVBench~\cite{yang2025svbench}, OVO-Bench~\cite{li2025ovo}, and StreamBench~\cite{xiong2025streaming} have introduced synchronous evaluation settings, yet they remain largely restricted to video question answering. As a result, they leave a broad range of practical online tasks unassessed, including live streaming narration, temporal grounding, and multi-turn interactive understanding.

To address these challenges, we present \textbf{LiveStarPro}, a proactive live streaming assistant for long-horizon video streams across diverse scenarios. LiveStarPro produces context-aware responses at semantically appropriate moments by combining adaptive streaming decoding with hierarchical memory management.
For \textbf{Challenge 1}, we establish a proactive response-silence paradigm coupling two synergistic innovations: a Streaming Verification Decoding \textbf{(SVeD)} mechanism that uses single-pass verification to determine the optimal response timing and suppress redundant outputs through strategic silence, and a stream-oriented training strategy built on Streaming Causal Attention Masks \textbf{(SCAM)} that aligns progressively revealed frames with their captions to instill the incremental video-language alignment SVeD requires.
For \textbf{Challenge 2}, we propose a Tree-Structured Hierarchical Memory \textbf{(TSHM)}: a Peak-End strategy distills the active context by prioritizing salient keyframes via perplexity verification, while evicted history is offloaded into a recursive event tree that attaches new events as children of semantically similar nodes, thereby modeling the temporal progression and causal dependencies of the stream; when a response is triggered, a context-aware retrieval mechanism re-injects relevant event chains of visual tokens and captions to support long-term reasoning.
Finally, to mitigate the data limitations, we introduce \textbf{OmniStarPro}, a comprehensive dataset for training and benchmarking that encompasses diverse real-world scenarios and evaluation tasks for online video understanding, and that augments the short-horizon setting with a long-form partition dedicated to long-term memory recall.
Extensive experiments across three benchmarks demonstrate that LiveStarPro attains state-of-the-art results for online video understanding.

The preliminary version of our work has been published in the proceedings of the Conference on Neural Information Processing Systems (NeurIPS) 2025~\cite{yang2026livestar}. This journal version extends it as follows. (1) Whereas the preliminary version targets instantaneous response timing over minute-scale streams, this version advances toward long-horizon online understanding, where an always-on assistant must process effectively unbounded hour-scale streams without succumbing to catastrophic forgetting, coupling proactive response timing with structured long-term memory for coherence over prolonged interactions. (2) We substantially extend the memory design: the Peak-End compression is reformulated as the Short-Term Working Memory of a more complete Tree-Structured Hierarchical Memory, which adds a Long-Term Retrieval Memory organized as a Recursive Event Tree, a memory-augmented generation mechanism based on hierarchical beam descent, and a theoretical analysis establishing bounded active memory and sublinear retrieval over unbounded streams. (3) We extend OmniStar to OmniStarPro by curating a new long-form partition (OmniStarPro-Long) of videos ranging from ten minutes to beyond one hour and three memory-centric tasks that probe recall of evicted historical content, in addition to the original 15 real-world scenarios and 5 short-horizon tasks. (4) We additionally update the experiments with evaluations on long-term recall, ablations of memory compression and retrieval, and efficiency analyses that raise the average semantic-correctness improvement from 19.5\% to 28.9\% over existing online Video-LLMs, expand the related work in Section~\ref{sec:related_work}, and publicly release our code, dataset, and resources at \url{https://github.com/sotayang/LiveStarPro}.

Our main contributions can be summarized as follows:

\begin{itemize}
    \item We present \textbf{LiveStarPro}, a proactive live streaming assistant that conducts continual real-time comprehension over long-horizon streams, sustains coherent contextual reasoning across diverse online tasks, and emits responses only at semantically appropriate moments.

    \item We propose Streaming Verification Decoding (\textbf{SVeD}), a novel inference framework that determines the optimal response timing through a single forward-pass verification. It decouples silence determination from vocabulary generation, thereby avoiding the pitfalls of EOS-based approaches while preserving real-time responsiveness.

    \item We design a training strategy based on Streaming Causal Attention Masks (\textbf{SCAM}) that, through interleaved frame-caption sequences, trains LiveStarPro to incrementally align variable-length video inputs with linguistic outputs, supporting the dynamic verification logic of SVeD.

    \item We introduce a Tree-Structured Hierarchical Memory (\textbf{TSHM}) for long-horizon streaming video, coupling Peak-End compression for the active window with a Recursive Event Tree for long-term storage, enabling the model to retrieve coherent event chains and reuse historical visual-textual information evicted from the active context.

    \item We construct \textbf{OmniStarPro}, a comprehensive dataset spanning 15 real-world scenarios and 5 short-horizon tasks, further contributing a long-form partition with 3 memory-centric tasks for long-term recall. Extensive experiments demonstrate the state-of-the-art performance of LiveStarPro, with an average improvement of 28.9\% in semantic correctness and an 18.2\% reduction in timing difference relative to existing online Video-LLMs.
\end{itemize}

\section{Related Work}
\label{sec:related_work}
\subsection{Video Large Language Models}
The steady maturation of Large Language Models (LLMs)~\cite{touvron2023llama,team2023gemini,achiam2023gpt,ouyang2022training,radford2018improving, wang2025video, zhou2024object} has propelled parallel progress in Video-LLMs~\cite{ataallah2024minigpt4, maaz2023video,li2023videochat,yang2023vid2seq,wang2022internvideo}, enabling them to address demanding tasks such as video captioning~\cite{chen2024sharegpt4video,xu2024pllava,islam2024video}, question answering~\cite{ko2023large,li2024mvbench,maaz2024videogpt+, 9770842,10214041,11506215}, and temporal grounding~\cite{guo2025vtg,xu2024vtg,wang2024hawkeye} within offline settings.
Representative open-source systems, such as LLaVA-NeXT-Video~\cite{zhang2024llavanext-video}, Video-LLaVA~\cite{lin2023video}, and VILA~\cite{lin2024vila}, together with closed-source systems such as GPT-4o~\cite{achiam2023gpt} and Gemini 1.5 Pro~\cite{reid2024gemini}, generally treat video as a pre-recorded and finite file.
Yet seamless human-computer interaction demands capabilities beyond static video analysis: an effective assistant should process real-time streams while exercising autonomous temporal decision-making, responding at contextually appropriate moments without explicit user prompts~\cite{chen2024videollm,qian2024streaming}.
Existing Video-LLMs~\cite{10721284,10670217,10815073,10839067} remain constrained by the dynamic nature of continuous streams and often lack the flexibility to alternate between passive observation and active response.

\subsection{Online Video Understanding}
Driven by real-time applications such as live streaming~\cite{gao2023livechat} and wearable devices~\cite{grauman2022ego4d}, recent work investigates architectures for online video understanding~\cite{qian2024streaming, chen2024videollm, ding2025streammind, yang2026don, zhang2026querystream}, where systems must process frames incrementally and decide \textit{when} to respond.
VideoLLM-online~\cite{chen2024videollm} pioneered a streaming EOS (End-Of-Sequence) token mechanism to mark silence intervals, later refined for efficiency by VideoLLM-MoD~\cite{wu2024videollm} and LION-FS~\cite{li2025lion}.
As StreamMind~\cite{ding2025streammind} notes, however, EOS reliance is structurally flawed: the imbalance between silent and active frames biases training, while the semantic conflict between rich visual inputs and the meaningless EOS token erodes visual-language alignment~\cite{zhu2023languagebind}.
Reliable evaluation is equally critical. Offline benchmarks~\cite{li2024mvbench,mangalam2024egoschema,yu2019activitynet,li2020hero,patraucean2024perception,zadeh2019social,xu2017video,lei2018tvqa,xiao2021next,song2024moviechat,wang2024lvbench,jang2017tgif} probe abilities from action recognition to long-term reasoning~\cite{11146594,11359544,11430664,11535730,11202655}, yet present complete, pre-segmented videos~\cite{yang2026never} unlike unsegmented live streams~\cite{11168273,11097075,11184436}.
Recent streaming benchmarks such as SVBench~\cite{yang2025svbench}, OVO-Bench~\cite{li2025ovo}, and StreamBench~\cite{xiong2025streaming} evaluate models during synchronous playback, but rely on Video Question Answering alone and overlook tasks like continuous narration or real-time grounding.
Their scenario coverage is also narrow, since heavy reliance on Ego4D~\cite{grauman2022ego4d} restricts evaluation primarily to first-person perspectives.
To support a more holistic evaluation, we introduce a comprehensive dataset and benchmark tailored to online agents that encompasses diverse real-world scenarios together with a synergistic set of streaming tasks.

\subsection{Memory Management for Long-Form Video}
A central bottleneck in processing continuous video streams is reconciling the finite context window of LLMs with arbitrarily long visual inputs. Early studies pursued token reduction through techniques such as spatial-temporal pooling~\cite{zhou2024streaming, wang2022internvideo} and token merging~\cite{ song2024moviechat}. More elaborate strategies were subsequently developed to capture extended temporal dependencies. Within streaming settings, VideoStreaming~\cite{qian2024streaming} propagates memory from earlier clips to guide the encoding of the current clip, while StreamChat~\cite{xiong2025streaming} and TimeChat-Online~\cite{yao2025timechat} employ hierarchical memory or differential token dropping to filter redundant information on the fly. For offline long-form analysis, MovieChat~\cite{song2024moviechat}, MA-LMM~\cite{he2024ma}, and VideoLLaMB~\cite{wang2025videollamb} rely on external memory banks that store and retrieve compressed historical features. Most of these mechanisms, however, are tailored to offline processing in which global context is fully accessible, or they apply uniform compression that does not distinguish salient moments from redundant ones in a live stream. Moreover, the external memory banks adopted by these methods are organized as flat collections that require an exhaustive scan with a retrieval cost that grows linearly with the number of stored units and a footprint that grows linearly with the stream duration. Drawing on the Peak-End rule~\cite{kahneman2000evaluation} from cognitive psychology, we introduce a Tree-Structured Hierarchical Memory that both compresses the active context by retaining salient frames and organizes evicted history into a structured retrieval bank~\cite{yang2024ldre,yang2024semantic} with sublinear expected retrieval, thereby granting access to long-term dependencies in infinite streams without saturating the context window.

\begin{figure*}[t]
\centering
\includegraphics[width=\textwidth]{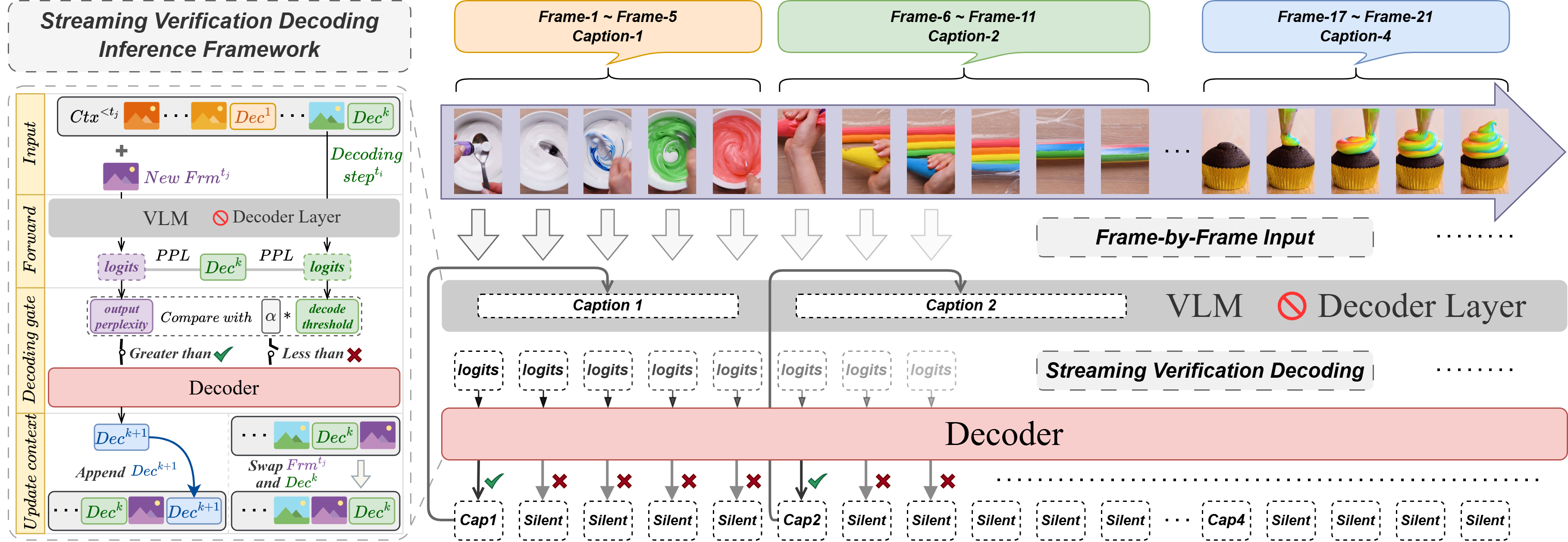}
\caption{\textbf{Overview of the streaming verification decoding (SVeD) inference framework:}
A dynamic response-silence decoding framework designed to determine optimal response timing for online video understanding.}
\label{fig:framework}
\vspace{-1mm}
\end{figure*}

\section{The LiveStarPro Framework}

\subsection{Overview}
\label{sec:overview}
To advance real-time online video understanding, we introduce \textbf{LiveStarPro}, a proactive live streaming assistant for long-horizon streams. As illustrated in Fig.~\ref{fig:overview}, the architecture comprises three sub-modules that jointly resolve response timing, streaming alignment, and long-horizon memory management. \textit{Streaming Verification Decoding (SVeD)} is a dynamic inference mechanism that autonomously identifies the appropriate response moment through a lightweight single-pass perplexity verification, removing the dependency on explicit silence tokens and delivering proactive context-aware responses with minimal latency. \textit{Streaming Causal Attention Masks (SCAM)} constitute an instruction-tuning paradigm tailored to the incremental behavior of SVeD; by imposing specialized causal masks on interleaved frame-caption sequences, it aligns evolving visual prefixes with linguistic semantics so that understanding is continually updated as new frames arrive. \textit{Tree-Structured Hierarchical Memory (TSHM)} manages memory for long-horizon streaming video, coupling a short-term working memory compressed through the Peak-End strategy with a long-term Recursive Event Tree, thereby enabling selective recall of historical event chains beyond the active context window and mitigating catastrophic forgetting throughout prolonged interactions.

\subsection{Inference Framework}
\label{sec:inference}

A central challenge in online video understanding lies in the trade-off between responsiveness and computational efficiency. Unlike offline models that operate over complete videos, a streaming assistant must autonomously decide \textit{when} to update its response as new frames arrive. A naive frame-by-frame decoding scheme guarantees high responsiveness, yet it incurs prohibitive latency together with substantial narrative redundancy. To overcome this difficulty, we devise an inference framework centered on Streaming Verification Decoding (SVeD) that shifts the paradigm from continuous generation toward efficient verification.

\subsubsection{Semantic Verification via Autoregressive Generation}
A straightforward way to determine response timing verifies the semantic validity of historical outputs against each incoming frame. Let $[Dec]^{t_i}$ denote the caption most recently emitted at timestamp $t_i$. For a new frame at time $t_j$, the model performs full autoregressive decoding to generate a candidate $[Dec]^{t_j}_{\mathrm{cand}}$, which is compared with the previous response to decide whether the content has evolved or remains redundant. Formally, the response policy is:
\begin{equation}
\text{Output}(t_j) =
\begin{cases}
\texttt{silent}, & \mathcal{S}([Dec]^{t_i}, [Dec]^{t_j}_{\mathrm{cand}}) \ge \tau \\
[Dec]^{t_j}_{\mathrm{cand}}, & \mathcal{S}([Dec]^{t_i}, [Dec]^{t_j}_{\mathrm{cand}}) < \tau
\end{cases}
\end{equation}
where $\mathcal{S}(\cdot,\cdot)$ is a semantic similarity function and $\tau$ a predefined threshold: a semantically consistent caption is suppressed as silence, otherwise the candidate is released.

This \textit{generate-then-compare} strategy, however, requires a complete autoregressive decoding for every frame before a decision, incurring $\mathcal{O}(L)$ latency per frame with sequence length $L$, which is impractical for low-latency real-time streaming.

\subsubsection{Streaming Verification Decoding}
To overcome this efficiency bottleneck, we propose \textbf{Streaming Verification Decoding (SVeD)}, which replaces the \textit{generate-then-compare} paradigm with a \textit{verify-then-generate} formulation. Rather than treating silence as a generation target, we regard it as a verification state.
SVeD introduces a lightweight \textit{decoding gate} that governs the transition between ``watching'' and ``speaking.'' The underlying mechanism requires only a single forward pass to verify the semantic validity of the existing caption with respect to the newly arriving visual frames. At any decoding step $t_i$, the perplexity (PPL) of the generated caption $[Dec]$ acts as a confidence measure:
\begin{equation}
\text{PPL}^{t_i}([Dec])=\sqrt[N]{\frac{1}{P([Dec] \ | \ [Ctx^{< t_i}],[Frm^{t_i}])}}
\end{equation}
where $N$ is the token length of $[Dec]$. For each subsequent incoming frame $[Frm^{t_j}]$, rather than autoregressively generating new tokens, SVeD simply re-evaluates the PPL of the \textit{previous} caption $[Dec]$ under the updated context.

Based on this metric, the gate operates on a threshold logic: if the verification perplexity exceeds a scaled reference value ($\text{PPL}^{t_j}([Dec]) > \alpha \cdot \text{PPL}^{t_i}([Dec])$), it indicates a significant divergence between the visual input and the current description, prompting the gate to activate and generate a fresh caption. Conversely, if the perplexity remains stable, the gate suppresses generation to avoid redundancy. In this suppression state, to ensure the context window reflects the passage of time, we perform a logical \textbf{Swap} operation: the existing caption $[Dec]$ is moved to the end of the context buffer $[Ctx]$, effectively extending its validity to the current timestamp. This mechanism, detailed in Algorithm~\ref{alg:SVeD}, ensures adaptive response timing and narrative coherence while significantly reducing computational overhead.

\begin{algorithm}[t]
\caption{Streaming Verification Decoding (SVeD)}
\label{alg:SVeD}
\begin{algorithmic}[1]
\REQUIRE Video frame stream $\{[Frm^{t}]\}_{t=1}^T$, Sensitivity threshold $\alpha$
\ENSURE Dynamically generated caption $[Dec]$
\STATE Initialize $[Dec],[Ctx] \leftarrow \emptyset$
\STATE Initialize reference timestamp $t_i \leftarrow 0$
\FOR{each incoming frame $[Frm^{t_j}]$}
    \STATE Append $[Frm^{t_j}]$ to $[Ctx]$
    \IF{$[Dec] \neq \emptyset$}
        \STATE Compute verification perplexity:
        \STATE $\text{PPL}^{t_j}([Dec]) = \sqrt[N]{1/P([Dec] \mid [Ctx^{\leq t_j}])}$
        \IF{$\text{PPL}^{t_j}([Dec]) > \alpha \cdot \text{PPL}^{t_i}([Dec])$}
            \STATE \textbf{Activate decoding:}
            \STATE Generate new tokens $[Dec]_{\text{new}}$ using $[Ctx^{\leq t_j}]$
            \STATE Update $[Dec] \leftarrow [Dec]_{\text{new}}$
            \STATE Append $[Dec]$ to $[Ctx]$
            \STATE $t_i \leftarrow t_j$ \COMMENT{Update reference timestamp}
        \ELSE
            \STATE \textbf{Suppress:} Swap the last two elements in $[Ctx]$
            \COMMENT{Move $[Dec]$ to the end}
        \ENDIF
    \ELSE
        \STATE Perform initial decoding to generate $[Dec]$
        \STATE Append $[Dec]$ to $[Ctx]$
        \STATE $t_i \leftarrow t_j$
    \ENDIF
\ENDFOR
\end{algorithmic}
\end{algorithm}

\subsubsection{Streaming Key-Value Cache}
An efficient implementation of SVeD must accommodate frequent context updates, including appends and swaps, without recurring computation. Conventional LLM caching mechanisms are largely static and therefore unable to cope with the dynamic sequence modifications that our framework induces. We accordingly devise an effective \textbf{Streaming Key-Value (KV) Cache} with a dual-level organization that maintains an \textit{intra-dialogue KV cache} for frame-level processing and an \textit{inter-dialogue streaming cache} for long-term context across conversations.
This design meets two critical requirements: it preserves cache sequence integrity under the logical swap operations of SVeD, removing recomputation of historical representations when the caption position changes; and it accommodates the dynamic length adaptation of Peak-End memory compression, permitting strategic pruning of redundant tokens while retaining temporal coherence. As reported in Tab.~\ref{abl:strategies}, this strategy accelerates inference by \textbf{1.58$\times$} over configurations without KV caching while incurring negligible performance loss, rendering SVeD highly suitable for online deployment.

\subsection{Training Strategy}
\label{sec:training}

Although the SVeD framework supplies an efficient means of determining contextually appropriate response timing, its effectiveness depends on the model's capacity to accurately estimate the probability of a caption under a progressively evolving visual context. Foundation models pre-trained on static image-text pairs typically lack the temporal granularity that the incremental verification logic of inference requires. To close this gap, we devise a stream-oriented training strategy centered on \textbf{Streaming Causal Attention Masks (SCAM)} that reformulates the training objective to match the dynamic inference requirements of LiveStarPro.

\subsubsection{Streaming Video-Language Alignment}
Existing Video-LLMs generally build upon foundation models pre-trained on static image-text pairs~\cite{chen2024expanding,Qwen2VL,liu2023visual}. Such models commonly optimize a static alignment objective:
\begin{equation}
\max P([Txt_i] \ | \ [Img_i]/[Vid_i]),
\end{equation}
which proves ill-suited to online scenarios in which the visual context accumulates incrementally and demands dynamic alignment with the linguistic outputs. To resolve this difficulty, we reformulate the training objective so as to model the probability of the current semantic description conditioned on the evolving historical context:
\begin{equation}
\max P([Txt^{k}] \ | \ [Ctx^{< t_i}],[Frm^{t_i}]),\ \forall t_i\in C_k,
\end{equation}
where $[Frm^{t_i}]$ represents the frame at timestamp $t_i$, and $[Ctx^{< t_i}]$ denotes the accumulated multimodal history. Here, $C_k$ defines a \textbf{semantic clip}, namely a continuous sequence of frames that share the same semantic event description $[Txt^{k}]$.
Crucially, this objective departs in a fundamental manner from EOS-based approaches~\cite{chen2024videollm, wu2024videollm} that compel the model to predict a silence token (i.e., $\max P(\text{EOS} \ | \dots)$) for non-response frames. By circumventing this trivial mapping, LiveStarPro sustains a consistent focus on meaningful visual-linguistic correlations and thereby furnishes a robust probability distribution for the verification stage of SVeD.

\subsubsection{Interleaved Frame-Caption Sequences}
To support frame-by-frame processing throughout training, we organize the data as \textbf{Interleaved Frame-Caption Sequences}. This format reproduces the streaming inference procedure in which the model continually ingests visual inputs and refreshes its understanding.
Concretely, for a semantic clip $C_k$, each frame $[Frm^{t_i}]$ is paired with the corresponding caption $[Cap^{k}]$. Because multiple frames within the same event share identical semantics, naive repetition could induce overfitting. To counter this tendency, we adopt a stochastic caption sampling strategy: for every frame, a caption $[Cap^{k}_j]$ is drawn at random from a predefined pool of $M$ paraphrased variants, which encourages the model to learn robust semantic representations rather than to memorize specific string patterns.



\begin{figure}[t]
\centering
\includegraphics[width=0.9\columnwidth]{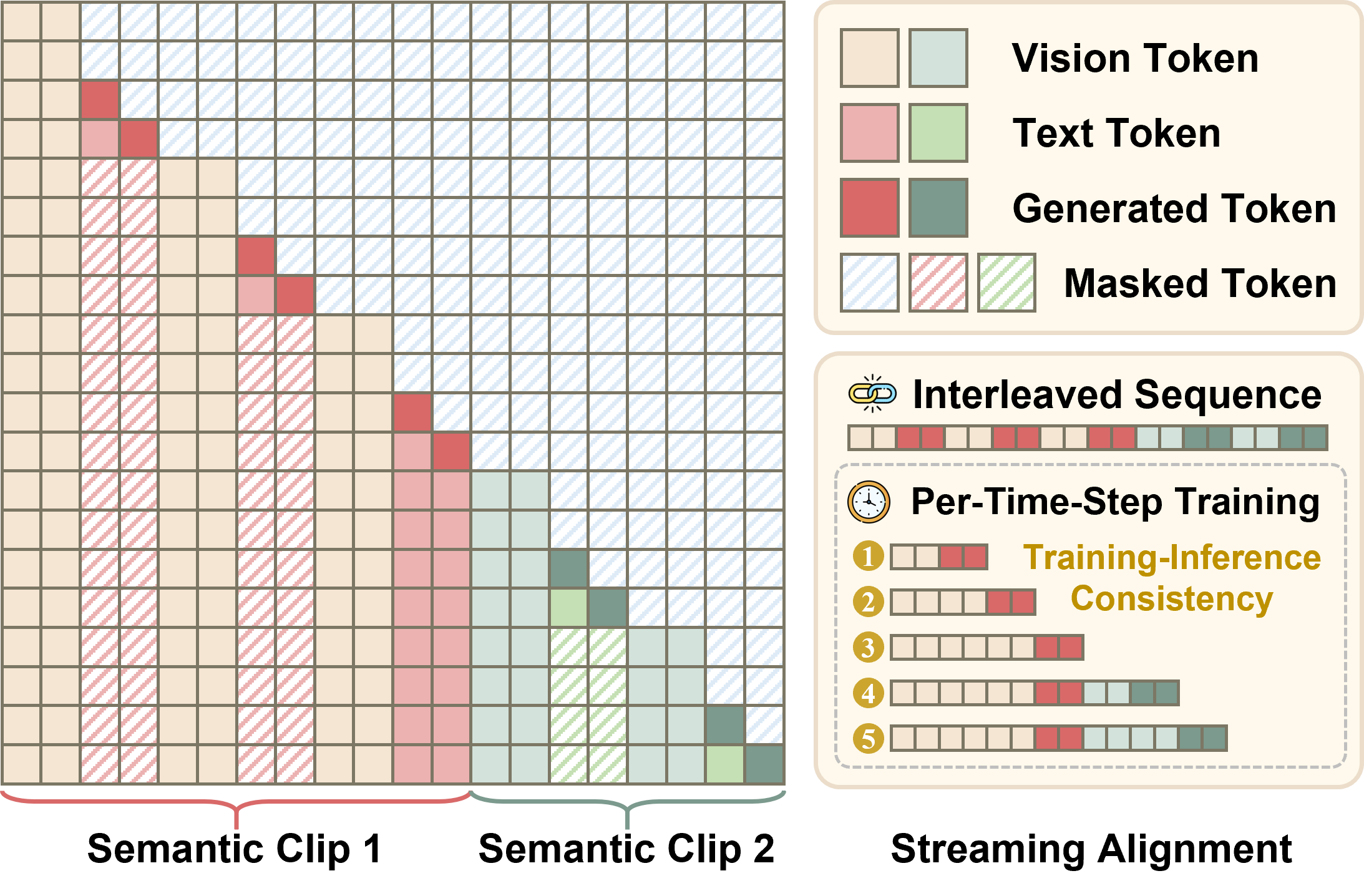}
\caption{\textbf{Overview of Streaming Causal Attention Masks (SCAM).} SCAM organizes frames and captions into interleaved sequences and performs progressive per-time-step training, masking preceding captions within each semantic clip to align training with streaming inference.
}
\label{fig:mask}
\end{figure}

\subsubsection{Streaming Causal Attention Masks}
Autoregressive training on these interleaved sequences raises distinctive challenges, particularly with respect to information leakage and context management. Since the frames within a single semantic clip $C_k$ share identical caption targets, a standard causal mask would permit the model to copy trivially the caption generated for a preceding frame in the same clip and consequently to disregard the current visual input.
To counter this behavior, we propose \textbf{Streaming Causal Attention Masks (SCAM)}, as illustrated in Fig.~\ref{fig:mask}. SCAM adapts the standard attention mechanism so as to enforce a selective visibility policy: it masks the attention weights that correspond to previously generated captions \textit{within the current semantic clip} $C_k$ and thereby compels the model to depend solely on the visual features of the current frame $[Frm^{t_i}]$ together with the accumulated history.
At the same time, in order to preserve narrative coherence across scene transitions, SCAM retains full visibility of the \textit{terminal captions} from all preceding semantic clips $\{C_1, \dots, C_{k-1}\}$. This explicit demarcation of semantic boundaries allows the model to reason about the event history while intra-event redundancy is suppressed.
The optimized objective under SCAM is defined as:
\begin{equation}
\max P([Cap^{k}_j] \ | \ [Ctx^{< t_i}\{Mask^{\leq t_i}\}],[Frm^{t_i}]),\ \forall t_i\in C_k.
\end{equation}
Here, $Mask^{\leq t_i}$ realizes these constraints in mathematical form and ensures that LiveStarPro learns to generate captions that are grounded in visual evidence rather than in linguistic repetition.

\subsection{Tree-Structured Hierarchical Memory (TSHM)}
\label{sec:memory}
\begin{figure}[t]
    \centering
    \includegraphics[width=1.0\linewidth]{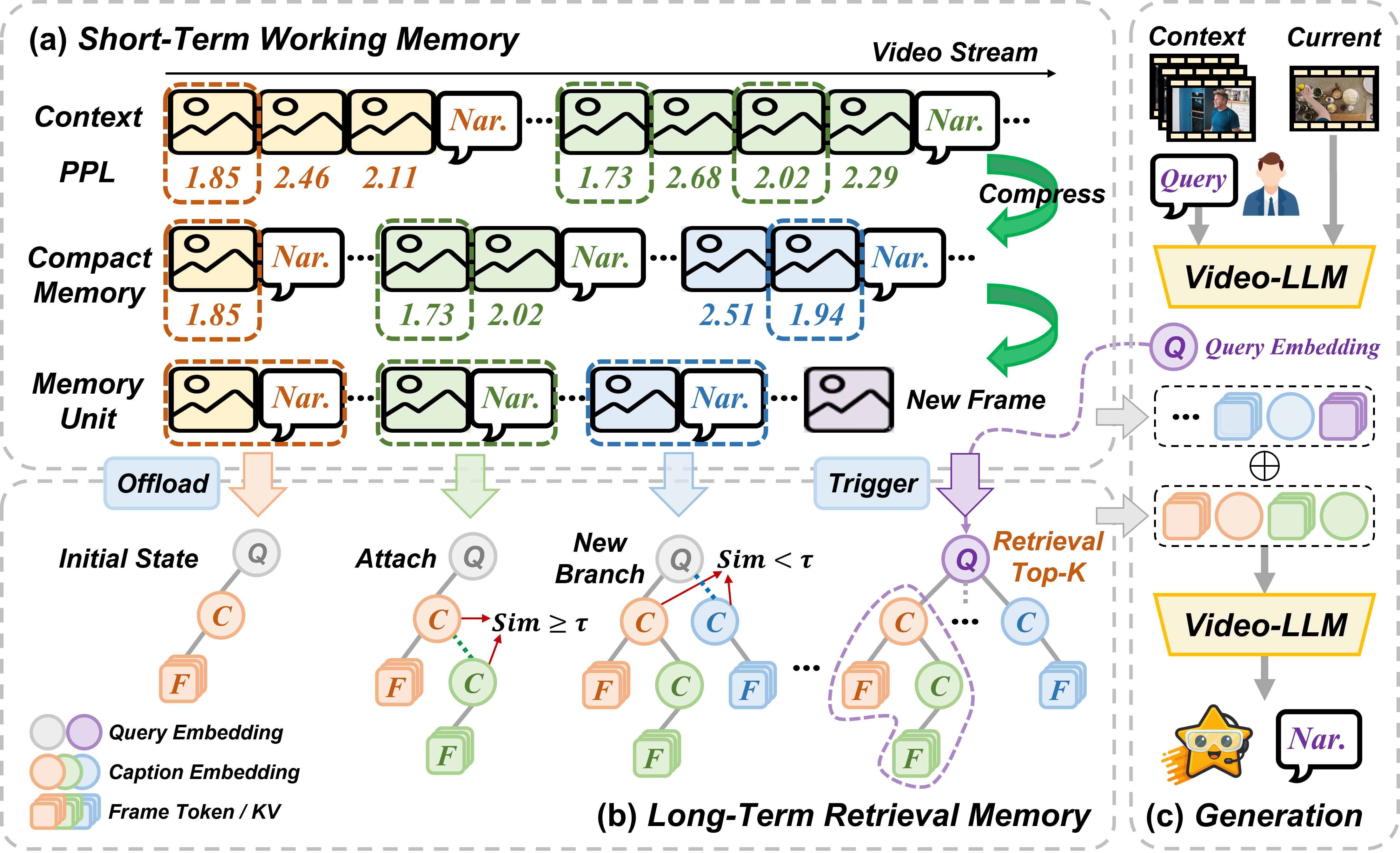}
    \caption{\textbf{Overview of Tree-Structured Hierarchical Memory (TSHM).} (a) Short-term frames are compressed via Peak-End rule, with evicted units offloaded to long-term storage. (b) The Recursive Event Tree organizes units by attaching similar events as children (Sim $\ge \tau$) or creating new branches. (c) Context-aware retrieval fetches relevant event chains to augment generation.}
    \label{fig:tshm}
\end{figure}

Although SVeD and SCAM endow LiveStarPro with robust capabilities for real-time inference and alignment, a critical deployment bottleneck arises from the memory accumulation and computational latency of long-duration video streams. Streaming inputs from live broadcasts, surveillance systems, and robotic cameras routinely extend well beyond the hour scale.
Whatever the context capacity, a perpetual stream will inevitably saturate the memory budget of the model. Conventional strategies such as First-In-First-Out (FIFO) or sliding window approaches are prone to catastrophic forgetting, since they discard historical events indiscriminately once observations drift past the fixed temporal horizon. To overcome these limitations, we propose \textbf{Tree-Structured Hierarchical Memory (TSHM)}. Inspired by human cognitive architectures, TSHM arranges memory into two complementary tiers: a high-resolution \textit{Short-Term Working Memory} that retains recent fine-grained details, and a compressed \textit{Long-Term Retrieval Memory} that preserves salient historical events in a structured and queryable format, as illustrated in Fig.~\ref{fig:tshm}.

\subsubsection{Short-Term Working Memory: Peak-End Compression}

The processing of modern streaming videos, which frequently span extended durations at high frame rates, imposes substantial computational challenges on long-horizon understanding. To mitigate this, we devise a memory compression mechanism that draws on the Peak-End Rule~\cite{kahneman2000evaluation}, according to which human memory predominantly retains salient moments together with summary-level representations of experiences. Our design distills the active context window, designated the \textbf{Short-Term Working Memory}, through the explicit modeling of these two complementary signals. Specifically, the \textbf{Peaks} are identified by keyframe selection that is grounded in semantic confidence. During the SVeD phase, the verification perplexity of each frame is computed as a semantic divergence score, defined as $S(t) = \text{PPL}^t([Dec])$. Frames with lower perplexity values signal a stronger semantic alignment with the ongoing description and are accordingly regarded as salient keyframes. In parallel, the \textbf{End} component is instantiated by the caption of each semantic clip, which functions as a temporal summary that captures the aggregated semantics of the entire event.

The accumulation of tokens within the short-term working memory is governed by a dynamic pruning strategy. Once the total token count attains the context budget $L_{max}$, a pruning cycle is launched independently for each semantic clip $C_k$. For every clip, a dynamic threshold $\tau_k$ is computed as the median divergence score of its frames, and only the following subset of frames is retained:
\begin{equation}
\mathcal{T}_{keep}^k = \{ t \in C_k \mid S(t) \leq \tau_k \}.
\end{equation}
This operation removes approximately 50\% of the frames with higher semantic divergence while consistently preserving the clip-level summary caption. As further frames arrive and additional pruning cycles are triggered, older semantic clips are progressively condensed into a compact representation consisting of a small number of salient peak frames together with their corresponding summary captions.
Should the short-term working memory remain saturated after all clips have been reduced to this compact form, the oldest semantic units are evicted and transferred to the long-term retrieval memory for persistent storage.

By prioritizing peak-level visual evidence and summary-level semantic representations, this compression strategy affords adaptive scaling for long-duration streams. As demonstrated in Tab.~\ref{abl:strategies}, LiveStarPro attains superior semantic correctness and lower timing difference relative to Uniform Dropout and FIFO-based forgetting strategies, which validates both its computational efficiency and its capacity to sustain coherent long-range narratives.

\subsubsection{Long-Term Retrieval Memory: Recursive Event Tree}
Even with the pruning mechanisms introduced above, an unbounded video stream will inevitably exceed the capacity of a finite context buffer. Rather than permanently discarding evicted memory units, we archive them in a tree-structured long-term memory that preserves temporal continuity and causal relationships among events. This memory is organized as a recursive event tree, where each node corresponds to a memory unit $U_i = \{c_i, v_i, \mathcal{E}_i, \mathcal{C}_i\}$. Here, $c_i$ denotes the event caption, $v_i$ represents the visual tokens of the corresponding peak frame, $\mathcal{E}_i$ is a semantic embedding used for indexing, and $\mathcal{C}_i$ stores the list of child nodes.
When a newly evicted unit $U_{new}$ is inserted into the long-term memory, its placement is determined by measuring semantic similarity between $\mathcal{E}_{new}$ and the embeddings of existing nodes. Specifically, we identify the node $U_{best}$ with the highest similarity score. If this score exceeds a predefined threshold $\sigma$, $U_{new}$ is attached as a child of $U_{best}$, indicating that the new event represents a refinement or continuation of an existing event thread. Otherwise, $U_{new}$ is initialized as a new root node, forming an independent event branch.

To ensure that higher-level nodes serve as effective semantic summaries of their corresponding event subtrees, the embedding of the parent node is updated upon each insertion using a momentum-based aggregation scheme:
\begin{equation}
\mathcal{E}_{parent} \leftarrow \text{Normalize}\left( (1-\beta) \cdot \mathcal{E}_{parent} + \beta \cdot \mathcal{E}_{child} \right),
\label{eq:momentum}
\end{equation}
where $\beta$ controls the update rate. This design allows parent embeddings to gradually evolve toward the semantic centroid of their descendant events, thereby supporting robust and discriminative retrieval.

\subsubsection{Memory-Augmented Generation via Retrieval}

The hierarchical organization of the long-term memory enables a unified retrieval-augmented generation framework that supports both explicit querying and implicit contextual reasoning. When the SVeD gate activates response generation, LiveStarPro constructs a query vector $q$ in accordance with the current task setting. In explicit question answering, $q$ is derived directly from the textual embedding of the user query, which permits the targeted retrieval of specific historical events or facts. By contrast, for streaming narration without an explicit user prompt, we formulate an implicit query from the aggregated visual embeddings of recent short-term frames, which captures the current visual context. This implicit visual query retrieves relevant historical events from the long-term memory and thereby allows the model to identify recurring entities or semantically related scenes and to maintain long-range narrative coherence.

Crucially, because each parent embedding is maintained as the semantic centroid of its subtree (Eq.~\ref{eq:momentum}), the query need not be scored against every stored unit. Instead, LiveStarPro performs a \textit{hierarchical beam descent}: starting from the set of root nodes, it scores only the immediate children of the current frontier by cosine similarity, retains the $k$ most similar nodes as the next frontier, and recurses toward the leaves. Formally, the frontier at depth $d$ is updated as
\begin{equation}
\mathcal{F}_{d+1} = \text{Top-}k_{\,U_i \in \text{Child}(\mathcal{F}_d)} \frac{q \cdot \mathcal{E}_i}{\|q\| \|\mathcal{E}_i\|},
\end{equation}
and the descent terminates at the leaf level, yielding the index set $\mathcal{I}$ of the retrieved units. Since only the children along the top-$k$ root-to-leaf paths are evaluated, this procedure inspects $O(k\,b\,\log_b n)$ nodes rather than all $n$ units, which realizes the sublinear retrieval cost analyzed in Sec.~\ref{sec:tshm_theory}. When the tree degenerates into isolated roots ($\sigma\!\to\!1$), the frontier spans the entire root level and the descent gracefully reduces to the exhaustive $O(n)$ scan of a flat index.

For each retrieved node, the associated event context is further gathered by traversing the corresponding tree paths, which include its parent and immediate children. The resulting memory set,
\[
\mathcal{M}_{retrieved} = \{ (c_j, v_j) \mid j \in \mathcal{I} \cup \text{Path}(\mathcal{I}) \},
\]
is injected back into the attention window during generation. By explicitly integrating retrieved historical captions and visual tokens with current observations, LiveStarPro effectively bridges long-term memory with ongoing perception, enabling coherent question answering together with temporally consistent streaming narration under a bounded context budget.

\subsubsection{Theoretical Analysis}
\label{sec:tshm_theory}
We characterize the memory footprint and retrieval cost of TSHM as a qualitative complexity argument under stated structural assumptions rather than a worst-case guarantee. Let $T$ denote the elapsed stream duration measured in semantic clips, and let $n$ denote the number of memory units in the long-term retrieval memory at query time.

\paragraph{Bounded active memory.}
The short-term working memory operates under a fixed token budget $L_{max}$. Whenever the budget is reached, the dynamic pruning rule retains the subset $\mathcal{T}_{keep}^k$ satisfying $S(t) \leq \tau_k$ with $\tau_k$ the per-clip median, removing about half of the frames of every clip while preserving the summary caption. Each clip is thus at most halved per pruning cycle, contributing at most $\lceil F_k / 2^{r} \rceil + 1$ units after $r$ cycles, where $F_k$ is its initial frame count. The active memory therefore never exceeds $L_{max}$ regardless of stream length, guaranteeing a constant per-step inference cost independent of $T$.

\paragraph{Expected logarithmic retrieval under balanced growth.}
A query traverses the recursive event tree from roots toward leaves. \emph{Under the assumption} that $\sigma$ and $\beta$ induce a bounded branching factor $b$ and balanced subtree growth, a tree storing $n$ units has height $h = O(\log_{b} n)$. Since retrieval evaluates the top-$k$ nodes along root-to-leaf paths and then expands the parent and immediate children of each, the expected number of similarity evaluations per query is $O(k\,b\,\log_{b} n)$, sublinear in the stored units, matching the moderate branching factor and shallow height observed on OmniStarPro-Long (Tab.~\ref{tab:abl:tree}). Balanced growth is not guaranteed: a highly skewed stream that collapses most events into one thread can degenerate the tree into a chain, while setting $\sigma$ so high that every unit forms an independent root reduces the structure to a flat index with linear $O(n)$ cost, which delineates the role of $\sigma$ in trading retrieval accuracy against efficiency.

\paragraph{Comparison with flat memory.}
A flat memory bank storing every evicted unit incurs $O(n)$ retrieval and a footprint growing linearly with $T$. TSHM instead bounds the active footprint by $L_{max}$, supports $O(\log n)$ expected retrieval under the balanced-growth assumption, and preserves causal event chains through parent and child traversal, which helps explain the graceful recall degradation of LiveStarPro as the memory span grows on OmniStarPro-Long.

\section{Dataset: OmniStarPro}

To enable rigorous evaluation and training of LiveStarPro, we construct \textbf{OmniStarPro}, a dataset for online video understanding that overcomes prior benchmarks with expert-annotated streams and a unified streaming protocol probing temporal perception and contextual awareness under strict causal constraints. It comprises two complementary partitions. \textbf{OmniStarPro-Live} inherits the five synergistic short-horizon tasks of the conference benchmark and emphasizes instantaneous response timing over minute-scale streams. \textbf{OmniStarPro-Long} is a newly curated collection of streams from ten minutes to beyond one hour, introducing three memory-centric tasks that probe recall of historical content after its eviction from the active context window. This long-form emphasis targets applications such as live streaming, surveillance, and cinematic tools.

\begin{figure*}[t]
\centering
\includegraphics[width=0.85\textwidth]{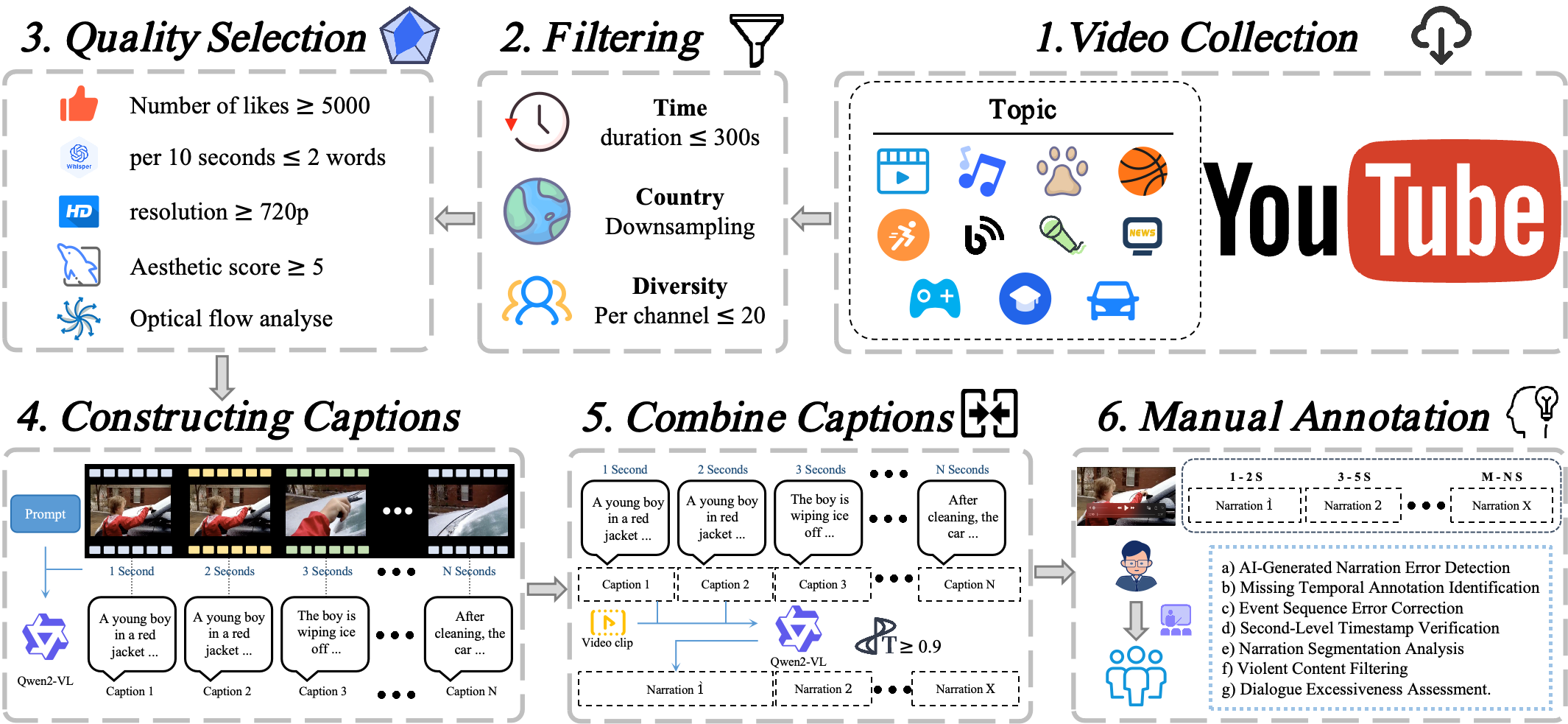}
\caption{\textbf{Overview of the pipeline of a rigorous multi-stage process.} Steps (1)-(3) involve data collection and preprocessing, and steps (4)-(6) involve constructing an online task dataset, using the OmniStarPro-RNG task as an example. Other online tasks are constructed in a similar manner.}
\label{fig:pipline}
\vspace{-1mm}
\end{figure*}

\subsection{Dataset Construction Pipeline}
\label{Dataset Pipeline}
The OmniStarPro construction follows a rigorous multi-stage process that guarantees multimodal consistency, content diversity, and real-time narration fluency. As depicted in Fig.~\ref{fig:pipline}, we take OmniStarPro-RNG as a running example and construct the remaining tasks analogously: steps (1)-(3) cover data collection and preprocessing shared across the five short-horizon tasks of the OmniStarPro-Live partition, whereas steps (4)-(6) assemble a specific online task dataset.

\subsubsection{Balanced Data Collection}
Data collection for the OmniStarPro-Live partition is initiated through the official API of YouTube, from which we gather 120,598 short videos ($\leq$ 6 minutes) together with the associated metadata.
We choose YouTube for its heterogeneous repository spanning broad categories and cultural distributions with detailed classification tags, concentrating on 15 real-world scenarios that encompass 46 specific categories (Fig.~\ref{fig:vis:video}).
To curb geographic bias, we apply stratified sampling conditioned on source countries via two mechanisms: downsampling overrepresented regions, and capping selection at 20 videos per channel, which prevents content homogenization while preserving platform authenticity. For the OmniStarPro-Long partition, we separately harvest 9,260 long-form videos exceeding ten minutes from the same scenarios as the raw pool for memory-centric curation.

\subsubsection{Multimodal Quality Filtering}
To safeguard cross-modal content quality, we implement a three-stage hierarchical filtering process. The social validation phase retains only videos with substantial audience engagement (more than 5,000 social interactions). For audio, we adopt OpenAI's Whisper\footnote{\url{https://github.com/openai/whisper}} speech recognition system under a strict lexical density constraint of at most two words per 10-second interval, since the evaluated Video-LLMs typically cannot process speech and excessively dense speech obscures their understanding of online videos.
The visual quality assessment comprises three sequential analyses. First, FFmpeg\footnote{\url{https://ffmpeg.org/}}-based resolution screening preserves only HD content ($\geq$720p). Subsequently, optical flow\footnote{\url{https://github.com/opencv/opencv/blob/3.4/samples/python/tutorial\_code/video/optical\_flow/optical\_flow.py}} analysis removes videos that exhibit either excessive motion blur ($\geq$15\% high-variance frames) or prolonged static segments ($\geq$85\% temporal occupancy within any 30-second window). Finally, an aesthetic assessment model\footnote{\url{https://github.com/hpcaitech/Open-Sora/tree/opensora/v1.3/tools/scoring}} is employed to exclude videos that score below 5/10 on visual composition metrics.
This multi-tiered filtering architecture progressively reduces the initial corpus from 120,598 candidate videos to a refined set of 21,544 high-quality videos.

\subsubsection{Temporal-Aware Frame Processing}
We operate on a per-second basis, generating an initial textual caption for each video second with Qwen2-VL~\cite{Qwen2VL}. To curb temporal redundancy while preserving narrative fluency, we introduce a dynamic segmentation algorithm that detects coherent story segments by analyzing semantic similarity between consecutive captions. The core of this approach relies on an adaptive thresholding mechanism ($\theta$ = 0.9) together with a LIFO (Last-In-First-Out) stack architecture. Specifically, timestamps are pushed onto the stack while the cosine similarity between successive captions is computed continuously. When adjacent captions exceed the semantic coherence threshold, the algorithm collapses them into a unified segment, which preserves temporal continuity while eliminating redundant descriptions. The final narration, produced by Qwen2-VL, combines two processes: (1) the merging of redundant captions within each coherent segment, and (2) the use of historical narrations to maintain contextual fluency between adjacent segments.

\subsubsection{Multidimensional Annotation}
\label{Annotation}
To guarantee high-quality annotations, we engaged a panel of 30 domain experts to perform multilayered annotation across seven critical dimensions: a) AI-Generated Narration Error Detection, b) Missing Temporal Annotation Identification, c) Event Sequence Error Correction, d) Second-Level Timestamp Verification, e) Narration Segmentation Analysis, f) Violent Content Filtering, and g) Dialogue Excessiveness Assessment. Through iterative consistency validation rounds, inter-annotator discrepancies were resolved and robust annotation agreement was established. This rigorous quality control yielded 20,137 validated video-text pairs with real-time narrations, an 87.8\% retention rate relative to the original dataset.

\subsubsection{Long-form Stream Curation}
The OmniStarPro-Long partition is assembled separately to support memory-centric evaluation over hour-scale content. Beginning from the raw pool of 9,260 long-form videos, we apply the same multimodal quality filtering and expert verification used for the short-horizon partition, which yields 2,108 high-quality long-form streams. These streams are stratified into three duration tiers: 1,396 of ten to thirty minutes, 331 of thirty to sixty minutes, and 381 beyond one hour, averaging 34.7 minutes. For each stream, per-second dense captions are produced and consolidated into coherent event segments through the dynamic semantic fusion procedure of StreamingCoT~\cite{hu2025streamingcot}, which furnishes reliable second-level temporal references. Building on these references, expert annotators compose 12,704 queries for the LMR, CDQ, and TBR tasks and record, for every query, the ground-truth timestamp of the supporting evidence together with the resulting memory span. This protocol ensures that a substantial proportion of the queries depend on content lying well beyond the active context window, thereby isolating the capacity for genuine long-term recall.

\subsection{Task Definitions}
To assess the capabilities of online agents comprehensively, we formulate two task families. The OmniStarPro-Live partition retains five short-horizon tasks that emulate real-world streaming interactions, whereas the OmniStarPro-Long partition contributes three memory-centric tasks that operate over long-form streams.

(1) \textit{Real-time Narration Generation (RNG)} requires the model to act as a live commentator producing coherent real-time descriptions of evolving content. Unlike offline captioning, it is penalized for latency and must decide when to speak or stay silent to avoid redundancy. (2) \textit{Online Temporal Grounding (OTG)} evaluates event localization within a continuous stream: given a query, the model must identify the start and end timestamps of the relevant segment as soon as it occurs, without access to future frames. (3) \textit{Frame-level Dense QA (FDQ)} probes fine-grained perception by querying the model at dense intervals about detailed visual attributes or actions in the current frame, assessing its capacity to sustain high-resolution awareness throughout the stream. (4) \textit{Contextual Online QA (COQ)} assesses short-term memory and causal reasoning through questions that depend on recent history, requiring the model to synthesize the current frame with the preceding context. (5) \textit{Multi-turn Interactive QA (MIQ)} emulates a continuous user-agent dialogue, challenging the model to handle coreference resolution and maintain a consistent persona over a long session while the background video continues to play.

(1) \textit{Long-range Memory Recall (LMR)} targets retrieval of factual details that appeared early in a long-form stream and have since been evicted from the active context window. Queried at a late timestamp about an attribute of a past entity or event, the model must recover it without the intervening frames. Because the query-evidence distance typically exceeds the active window, this task isolates genuine long-term recall rather than recent-context perception. (2) \textit{Cross-event Difference Query (CDQ)} compares two events separated by a long interval within the same stream: given a pair of widely spaced moments, the model must report the change in a salient attribute such as the count, state, or spatial arrangement of the entities. Requiring simultaneous retrieval and contrast of two distant memories, it probes access to multiple historical entries rather than an explicit logical chain. (3) \textit{Temporal Backtracking (TBR)} requires the model to locate the most recent past occurrence of a queried event and report its timestamp or an associated attribute. Unlike OTG, which localizes an event in the present, this task backtracks through evicted history to recover an event that has left the active window, exposing the catastrophic forgetting of fixed-window architectures.

\subsection{Statistics and Comparison}
OmniStarPro sets itself apart from existing benchmarks through its scale, its diversity, and its dedicated emphasis on online streaming constraints across a wide temporal spectrum.

\subsubsection{Scenario Diversity} 
\begin{figure*}[t]
\centering
\subfloat[]{
    \includegraphics[height=0.20\textheight]{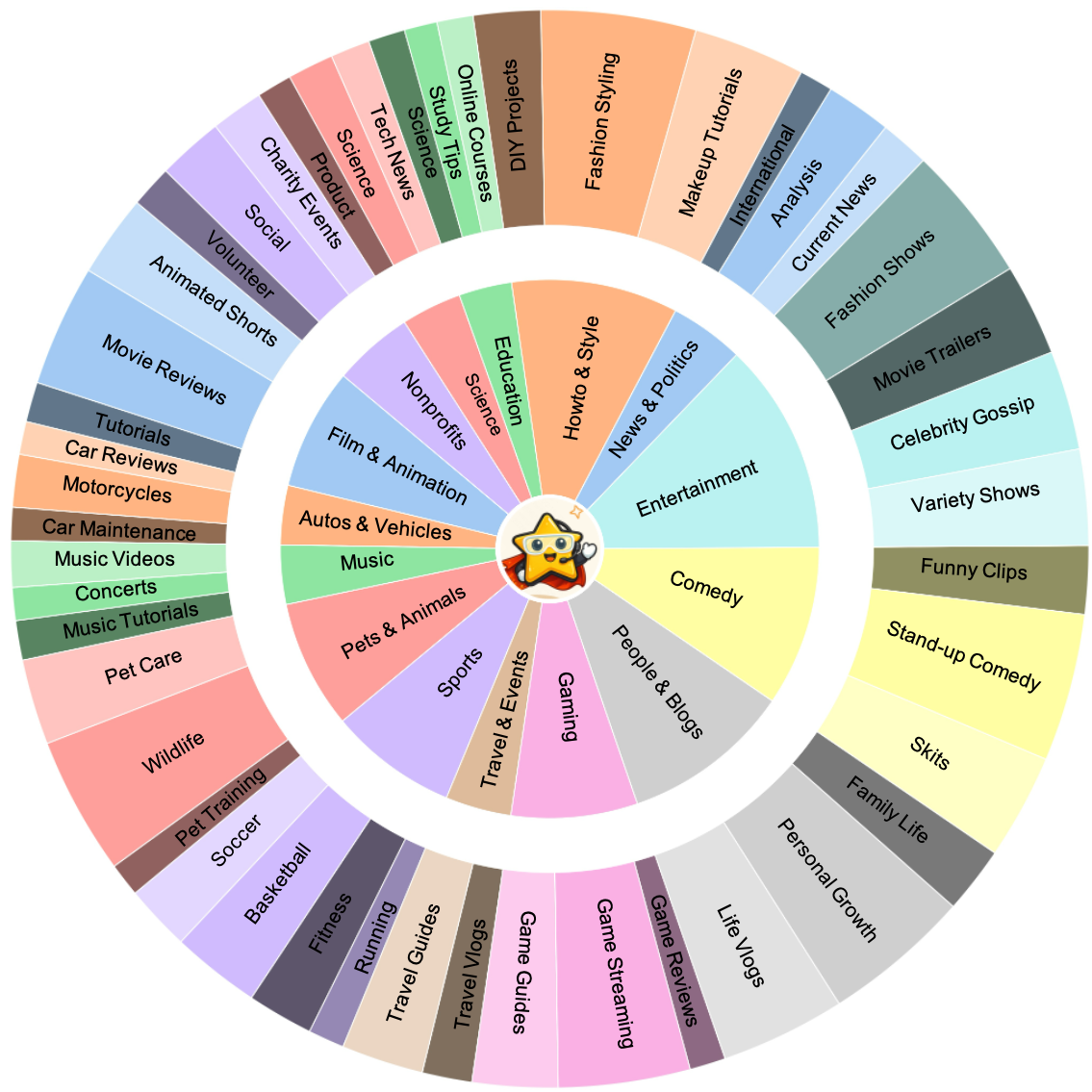}
    \label{fig:vis:video}
}
\hfil
\subfloat[]{
    \includegraphics[height=0.20\textheight]{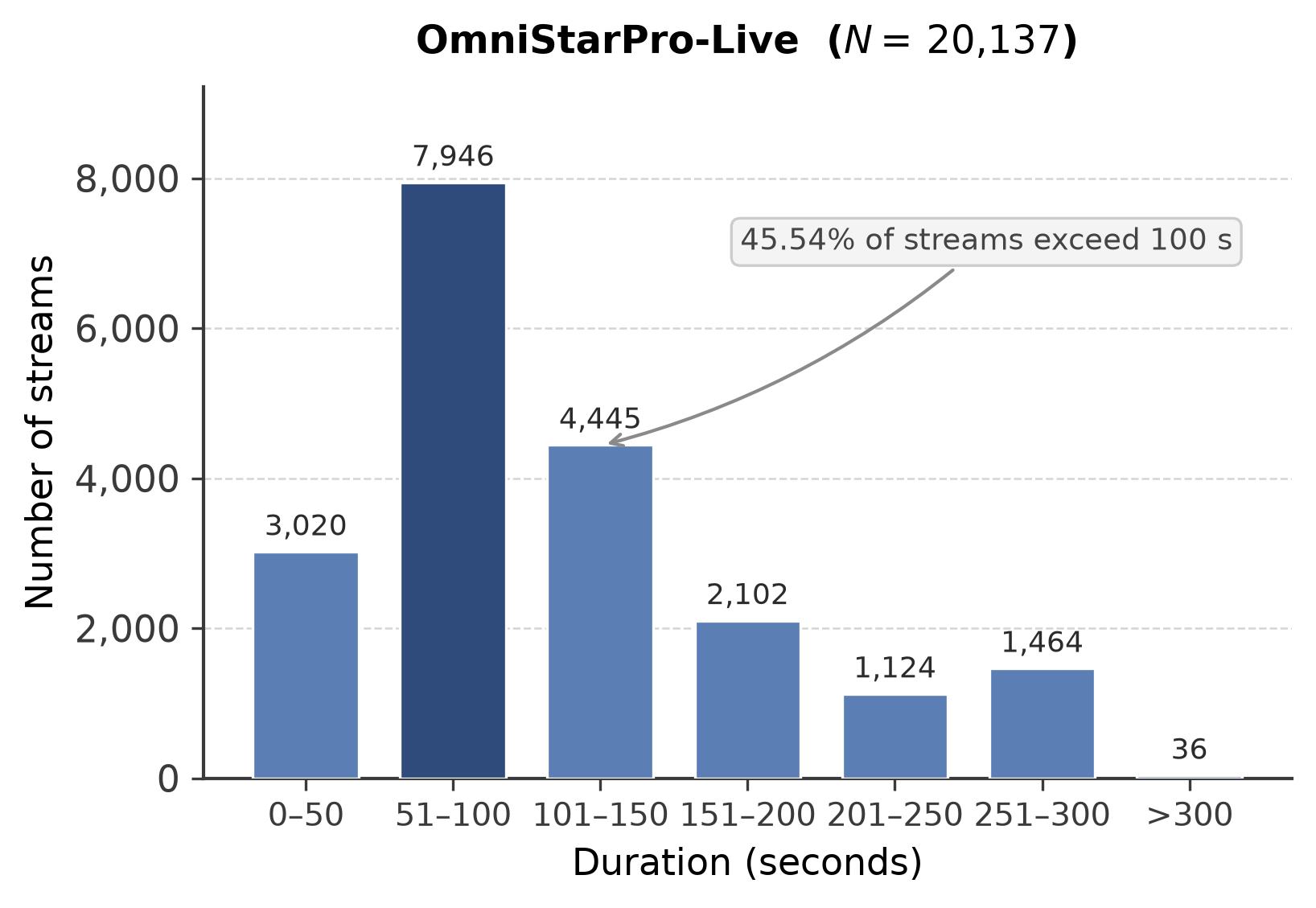}
    \label{fig:vis:length:live}
}
\hfil
\subfloat[]{
    \includegraphics[height=0.20\textheight]{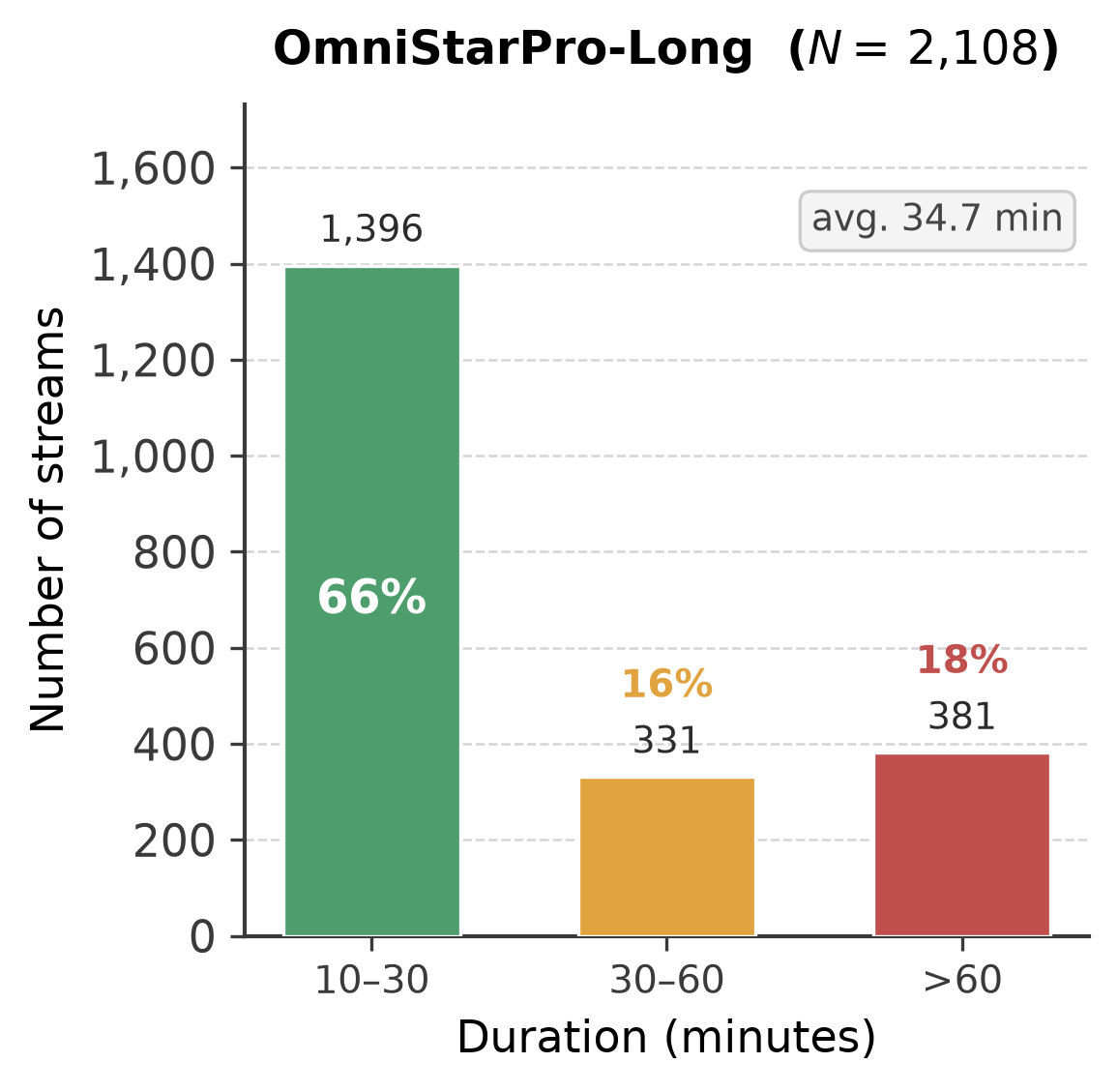}
    \label{fig:vis:length:long}
}
\caption{\textbf{Distributions of video data.} (a) Distribution of video categories across 15 real-world scenarios. (b) Duration distribution of the OmniStarPro-Live partition at the second level. (c) Duration distribution of the OmniStarPro-Long partition at the minute level.}
\vspace{-1mm}
\label{fig:visualization}
\end{figure*}

As illustrated in Fig.~\ref{fig:vis:video}, OmniStarPro covers 15 diverse real-world scenarios, including Travel \& Events, Sports, News \& Politics, and Gaming. Each scenario is further subdivided into 2 to 4 fine-grained categories by the native annotation system of YouTube, resulting in 46 specific categories in total.

\subsubsection{Video Length Distribution} Fig.~\ref{fig:vis:length:live} and Fig.~\ref{fig:vis:length:long} present the distribution of video durations across the two partitions. Within the OmniStarPro-Live partition, 45.54\% of the videos exceed 100 seconds in length, with the majority falling within the 51 to 150-second range. The OmniStarPro-Long partition extends this spectrum substantially, with 2,108 streams averaging 34.7 minutes (stratified as above). This coverage of hour-scale content confirms that OmniStarPro supports the joint evaluation of instantaneous response and long-term recall, far surpassing standard short-video benchmarks.

\subsubsection{Memory-Span Distribution} For the three memory-centric tasks, we additionally report the distribution of the memory span, defined as the temporal interval between a query and its supporting evidence. Across the 12,704 long-form queries, the memory span averages 18.6 minutes and reaches a maximum of 71.3 minutes, and 73.4\% of the queries exhibit spans that exceed the active context window of the model, which guarantees that the OmniStarPro-Long partition genuinely stresses long-term retrieval rather than recent-context perception.

\subsubsection{Annotation Scale} The OmniStarPro-Live partition contains 20,137 expert-annotated video streams that are partitioned into 19,137 training and 1,000 evaluation instances, with temporally dense annotations that average 14.5 QA pairs and 8.2 caption segments per video. The OmniStarPro-Long partition supplies an additional 2,108 long-form streams together with 12,704 memory-centric queries for the LMR, CDQ, and TBR tasks, each accompanied by a ground-truth evidence timestamp that supports span-aware evaluation. Together, the two partitions comprise 22,245 streams and ensure rigorous evaluation across both short-horizon and long-horizon online settings.

\section{Experiments}
\subsection{Experimental Setup}
\label{Experimental Setup}

\subsubsection{Datasets and Benchmarks}
We validate LiveStarPro across both offline and online settings. For real-world online evaluation, we adopt our OmniStarPro benchmark over the five short-horizon tasks and the three memory-centric long-form tasks. We further report results on SVBench~\cite{yang2025svbench} for streaming QA and build an Ego4D Narration Stream~\cite{grauman2022ego4d} benchmark from dense timestamped narrations to gauge egocentric understanding.
To endow LiveStarPro with streaming capabilities, we design a two-phase progressive training paradigm that draws on 83K samples.
\textbf{Phase I (Temporal Alignment Pretraining)} focuses on establishing frame-level semantic correspondences using 63K curated segments from ActivityNet Captions~\cite{yu2019activitynet} (9K selected from 20K raw samples), Shot2Story~\cite{han2023shot2story20k} (33K selected from 43K raw samples), Ego4D Narration Stream~\cite{grauman2022ego4d} (20K selected from 113K raw samples), and MVBench~\cite{li2024mvbench} (1K selected from 4K raw samples).
\textbf{Phase II (Multi-Task Online Adaptation)} utilizes 20K samples from OmniStarPro to specialize the model for the five online tasks via multi-objective alignment.

\subsubsection{Implementation Details}
LiveStarPro is built upon the InternVideo2.5 architecture~\cite{wang2025internvideo2,wang2022internvideo} and comprises an InternViT~\cite{chen2024expanding} vision encoder, an MLP projector, and an InternLM2.5-7B~\cite{cai2024internlm2} language model. InternViT extracts video frame embeddings at 1-4 FPS, with each frame represented by 16 tokens. For efficiency, we fine-tune the model under a static resolution strategy, which processes multi-minute content within an 8K-token context window. Full fine-tuning ran on 8$\times$ NVIDIA A800 GPUs.

\subsubsection{Settings.} During training, we trained the models for 1 epoch with a learning rate of $4 \times 10^{-5}$ using the AdamW optimizer ($\beta_1 = 0.9, \beta_2 = 0.999$, weight decay = 0.05). We utilized a per-device batch size of 1 and gradient accumulation over 4 steps to achieve an effective global batch size of 32. A cosine learning rate scheduling was adopted with a warmup ratio of 0.03. Input frames were uniformly resized to $448 \times 448$, with a patch downsampling ratio of 0.5. The vision encoder was frozen during training, while the MLP projector and language model components were fully updated. Each training sequence contains up to 8192 tokens, consisting of interleaved frame and language tokens following the InternVL2.5 conversational template. We optimized the model using the standard autoregressive cross-entropy loss computed over the language tokens; critically, the loss was computed only on assistant response tokens, and inter-frame language segments were excluded via our SCAM strategy.
For inference, the tunable scaling factor $\alpha$ in SVeD was set to 1.03 by default. Regarding the TSHM memory module, the pruning window $W$ in the short-term working memory was set to 60 frames, and the size of the paraphrased caption pool for streaming video-language alignment was set to $M=1$ by default to ensure optimal temporal alignment.

\begin{table*}[t]
\footnotesize
\centering
\caption{Quantitative comparison on the OmniStarPro-RNG task under offline and online settings. PPL is provided for reference only due to vocabulary variations. ``-'' denotes test inapplicability.}
\label{tab:rng_evaluation}

\setlength{\tabcolsep}{3pt}
\renewcommand{\arraystretch}{0.85}

\resizebox{0.9\textwidth}{!}{
\begin{tabular}{@{}lccccccccc@{}}
\toprule
\multirow{2}{*}{\textbf{Method}}
& \multicolumn{4}{c}{\textbf{Offline Evaluation}}
& \multicolumn{5}{c}{\textbf{Online Evaluation}} \\
\cmidrule(lr){2-5} \cmidrule(lr){6-10}
& \textbf{PPL$\downarrow$}
& \textbf{TokAcc$\uparrow$}
& \textbf{SemCor$\uparrow$}
& \textbf{SumFluen$\uparrow$}
& \textbf{TimDiff$\downarrow$}
& \textbf{TimRedun$\downarrow$}
& \textbf{TimCover$\uparrow$}
& \textbf{SemCor$\uparrow$}
& \textbf{SumFluen$\uparrow$} \\
\midrule
Human
& -- & -- & 6.73 & 7.17
& 1.08 & 1.24 & 0.84 & 6.09 & 6.81 \\
\midrule
\multicolumn{10}{c}{\textbf{Offline Video-LLMs / LVLMs}} \\
\midrule
GPT-4V~\cite{yang2023dawn}
& -- & -- & 4.97 & 5.37
& -- & -- & -- & -- & -- \\
GPT-4o~\cite{achiam2023gpt}
& -- & -- & 5.03 & 5.45
& -- & -- & -- & -- & -- \\
LLaVA-Video~\cite{zhang2024llavanext-video}
& 12.42 & 0.53 & 3.40 & 2.88
& -- & -- & -- & -- & -- \\
InternVideo2.5~\cite{wang2025internvideo2}
& 6.91 & 0.56 & 4.32 & 3.61
& -- & -- & -- & -- & -- \\
InternVL2.5~\cite{chen2024expanding}
& 9.81 & 0.51 & 3.40 & 2.94
& -- & -- & -- & -- & -- \\
MiniCPM-V 2.6~\cite{yao2024minicpm}
& 9.46 & 0.57 & 4.34 & 4.13
& -- & -- & -- & -- & -- \\
Qwen2.5-VL~\cite{bai2025qwen2}
& 13.80 & 0.59 & 4.42 & 4.24
& -- & -- & -- & -- & -- \\
\midrule
\multicolumn{10}{c}{\textbf{Online Assistants}} \\
\midrule
VideoLLM-online~\cite{chen2024videollm}
& 9.73 & 0.49 & 3.01 & 0.69
& 2.67 & 2.15 & 0.80 & 1.68 & 0.59 \\
VideoLLM-MoD~\cite{wu2024videollm}
& 9.93 & 0.48 & 2.89 & 0.65
& 2.54 & 2.49 & \textbf{0.90} & 1.66 & 0.55 \\
MMDuet~\cite{wang2024videollm}
& 5.69 & 0.60 & 4.29 & 3.40
& 2.32 & \textbf{0.62} & 0.51 & 1.93 & 2.69 \\
\rowcolor{gray!15}
LiveStar
& \textbf{5.14} & 0.62 & 4.62 & 4.55
& 1.91 & 0.95 & 0.71 & 3.19 & 4.25 \\
\rowcolor{gray!15}
LiveStarPro
& 5.21 & \textbf{0.65} & \textbf{4.76} & \textbf{4.64}
& \textbf{1.89} & 1.01 & 0.78 & \textbf{3.27} & \textbf{4.41} \\
\bottomrule
\end{tabular}
\vspace{-4mm}
}

\end{table*}

\begin{table}[t]
\footnotesize
\centering
\caption{Evaluation results of online Video-LLMs on OmniStarPro. ``--'' in OTG means no generation needed for scoring; ``--'' in COQ and MIQ indicates real-time QA.}
\label{tab:OmniStar}

\setlength{\tabcolsep}{3pt}
\renewcommand{\arraystretch}{1}

\begin{tabular}{@{}lccccc c@{}}
\toprule
\multirow{2}{*}{\textbf{Method}}
& \multicolumn{5}{c}{\textbf{Online Evaluation (SemCor$\uparrow$/TimDiff$\downarrow$)}}
& \multirow{2}{*}{\shortstack{\textbf{FPS$\uparrow$} \\ (\textbf{5 min})}} \\
\cmidrule(lr){2-6}
& \textbf{RNG}
& \textbf{OTG}
& \textbf{FDQ}
& \textbf{COQ}
& \textbf{MIQ}
&  \\
\midrule
Human
& 6.09/1.08
& --/1.81
& 9.12/1.01
& 7.96/--
& 7.83/--
& -- \\
\midrule
VideoLLM-online
& 1.68/2.67
& --/9.69
& 2.35/2.15
& 4.01/--
& 3.83/--
& 3.37 \\
VideoLLM-MoD
& 1.66/2.54
& --/9.83
& 2.11/2.23
& 3.99/--
& 3.75/--
& 3.41 \\
MMDuet
& 1.93/2.32
& --/4.42
& 4.78/2.65
& 5.71/--
& 5.62/--
& 0.91 \\
\rowcolor{gray!15}
LiveStar
& 3.19/1.91
& \textbf{--/3.57}
& 6.44/1.80
& 5.85/--
& 5.78/--
& 3.82 \\
\rowcolor{gray!15}
LiveStarPro
& \textbf{3.27/1.89}
& --/3.61
& \textbf{6.61/1.77}
& \textbf{5.97/--}
& \textbf{5.81/--}
& \textbf{3.96} \\
\bottomrule
\end{tabular}

\end{table}

\subsection{Online Experiments}
To approximate real-world online conditions, we conduct inference-time evaluations of Video-LLMs on the OmniStarPro test set. Unlike the Ego4D Narration Stream benchmark, which runs offline for lack of inference-result scoring, and semi-online benchmarks such as SVBench that rely on fixed decoding timestamps, OmniStarPro lets Video-LLMs autonomously decide when to respond or remain silent across five tasks in a fully online setting. This setting assesses both the temporal accuracy and the semantic consistency of responses against the ground truth.

\textbf{Evaluation Metrics.}
We adopt the following metrics to assess model performance as an online video assistant.
For each video, we denote the set of ground-truth semantic clips as $G=\{g_1,\dots,g_N\}$, where each clip $g_i$ is associated with a temporal interval $[t_{\text{start},i}, t_{\text{end},i}]$ and a ground-truth caption. The model produces a set of responses $R=\{r_1,\dots,r_M\}$, where each response $r_j$ is generated at time $t_{\text{resp},j}$.
For each semantic clip $g_i$, we define the set of matched responses as
\[
M_i = \{ r_j \in R \mid t_{\text{resp},j} \in [t_{\text{start},i}, t_{\text{end},i}] \}.
\]

\begin{itemize}
\item
\textbf{TimDiff (Timing Difference)} measures the temporal deviation between model-generated responses and ground-truth semantic clips. For each clip, missing responses are penalized by assigning the full clip duration, while multiple responses incur cumulative latency penalties. Formally, TimDiff is defined as
\[
\begin{aligned}
\text{TimDiff} = \frac{1}{N} \sum_{i=1}^N \Big(
& \mathbb{I}[|M_i| = 0] \cdot (t_{\text{end},i} - t_{\text{start},i}) \\
& + \mathbb{I}[|M_i| > 0] \cdot \sum_{r_j \in M_i}
(t_{\text{resp},j} - t_{\text{start},i})
\Big),
\end{aligned}
\]
where $\mathbb{I}[\cdot]$ denotes the indicator function. Lower TimDiff values indicate closer temporal alignment between responses and visual events.

\item
\textbf{TimRedun (Timing Redundancy)} evaluates response redundancy by measuring the deviation from the ideal case of producing exactly one response per semantic clip. It is computed as the average absolute difference between the number of generated responses and one across all clips:
\[
\text{TimRedun} = \frac{1}{N} \sum_{i=1}^N \big| |M_i| - 1 \big|.
\]

\item
\textbf{TimCover (Timing Coverage)} measures the proportion of semantic clips that receive at least one valid response. It is defined as
\[
\text{TimCover} = \frac{1}{N} \sum_{i=1}^N \mathbb{I}[|M_i| > 0],
\]
where a clip contributes a score of 1 if it contains at least one response and 0 otherwise.

\item
\textbf{SemCor (Semantic Correctness)} assesses semantic alignment between model responses and ground truth using GPT-4o. For each semantic clip $g_i$, if $|M_i| \geq 1$, we select the earliest response in $M_i$ as the representative output; otherwise, an empty response is assigned. GPT-4o scores the response--clip pair along three dimensions: (1) Semantic Accuracy, (2) Language Quality, and (3) Information Completeness, each on a 0--10 scale. The final SemCor score is obtained by averaging the three dimensions.

\item
\textbf{SumFluen (Summarize Fluency)} evaluates the holistic fluency and narrative quality of the model’s complete output sequence. All responses are concatenated in temporal order to form a model narrative, which is compared against the concatenated ground-truth narrative using GPT-4o. Evaluation is conducted along five dimensions: (1) Writing Logicality, (2) Language Fluency, (3) Writing Conciseness, (4) Semantic Consistency, and (5) Narrative Completeness.
\end{itemize}

\noindent\textbf{Validity of LLM-as-Judge.}
SemCor and SumFluen use GPT-4o as an automated judge~\cite{zheng2023judging,liu2023g}. Since all methods are scored with identical prompts and multi-dimensional rubrics, the judge applies a consistent scoring function; thus, although absolute scores may carry stylistic bias, the rank-order and relative improvements across methods remain reliable.

\textbf{Results.}
The right portion of Tab.~\ref{tab:rng_evaluation} reports the online evaluation results of different models on OmniStarPro-RNG, where each model follows its own response-silence strategy. The results show that LiveStarPro achieves both reduced response latency and improved semantic accuracy. Notably, VideoLLM-online and VideoLLM-MoD generate outputs for nearly every frame, resulting in the highest TimCover scores; however, this aggressive behavior leads to inferior performance on other metrics. Conversely, MMDuet produces substantially fewer responses, yielding the lowest TimRedun but at the expense of reduced performance on other evaluation criteria. These trends are further illustrated in Tab.~\ref{tab:OmniStar}, where LiveStarPro surpasses prior online Video-LLMs across the OmniStarPro tasks and improves over its conference predecessor LiveStar on most tasks, with comparable timing on OTG, while maintaining the fastest inference speed. Relative to the second-best prior online Video-LLM, it achieves a 28.9\% improvement in SemCor, an 18.2\% reduction in TimDiff, and a 16.1\% increase in FPS.

\textbf{Long-form Memory Evaluation.}
To assess long-term recall, we further evaluate online Video-LLMs on the OmniStarPro-Long partition across the three memory-centric tasks. Following the span-aware protocol, we report recall accuracy under three memory-span buckets, namely short ($<$10 minutes), medium (10--30 minutes), and long ($>$30 minutes), which respectively correspond to evidence within, near the boundary of, and well beyond the active context window. Here recall accuracy denotes the proportion of queries whose generated answer matches the ground-truth attribute or timestamp of the supporting evidence, judged by GPT-4o against the per-query ground truth, rather than a top-$k$ retrieval recall. As summarized in Tab.~\ref{tab:OmniStarLong}, all sliding-window baselines exhibit a pronounced degradation as the memory span grows, with accuracy that collapses toward chance once the supporting evidence is evicted from the active window. By contrast, LiveStarPro retains a substantially higher recall across all three buckets and degrades far more gracefully on the long bucket, which directly evidences the benefit of the hierarchical memory of TSHM for genuine long-term retrieval rather than recent-context perception.
These results also serve as a principled proxy for comparison with offline memory architectures such as MA-LMM~\cite{he2024ma}, MovieChat~\cite{song2024moviechat}, and VideoLLaMB~\cite{wang2025videollamb}: all three methods organize evicted features in flat external banks identical in structure to the flat $k$-NN baseline in Tab.~\ref{tab:abl:retrieval}, which obtains 21.3\% on the long bucket. The recursive event tree in TSHM achieves 37.2\% on the same partition, demonstrating a clear advantage over the flat-retrieval paradigm that these prior works rely on.

\begin{table}[t]
\footnotesize
\centering
\caption{Recall accuracy (\%) on the OmniStarPro-Long partition under three memory-span buckets. ``S'', ``M'', and ``L'' denote short ($<$10 min), medium (10--30 min), and long ($>$30 min) spans.}
\label{tab:OmniStarLong}
\setlength{\tabcolsep}{3pt}
\renewcommand{\arraystretch}{1}
\begin{tabular}{@{}lccc ccc ccc@{}}
\toprule
\multirow{2}{*}{\textbf{Method}}
& \multicolumn{3}{c}{\textbf{LMR}}
& \multicolumn{3}{c}{\textbf{CDQ}}
& \multicolumn{3}{c}{\textbf{TBR}} \\
\cmidrule(lr){2-4}\cmidrule(lr){5-7}\cmidrule(lr){8-10}
& \textbf{S} & \textbf{M} & \textbf{L}
& \textbf{S} & \textbf{M} & \textbf{L}
& \textbf{S} & \textbf{M} & \textbf{L} \\
\midrule
VideoLLM-online
& 41.2 & 18.6 & 6.4
& 33.7 & 14.1 & 5.2
& 38.5 & 16.0 & 5.8 \\
VideoLLM-MoD
& 42.0 & 19.3 & 6.9
& 34.5 & 14.8 & 5.6
& 39.1 & 16.7 & 6.1 \\
MMDuet
& 47.8 & 24.5 & 9.1
& 39.6 & 19.2 & 7.4
& 44.2 & 21.3 & 8.0 \\
\rowcolor{gray!15}
LiveStar
& 59.5 & 33.0 & 21.1
& 51.3 & 28.1 & 17.7
& 56.0 & 30.2 & 19.9 \\
\rowcolor{gray!15}
LiveStarPro
& \textbf{63.4} & \textbf{49.7} & \textbf{37.2}
& \textbf{55.1} & \textbf{42.8} & \textbf{31.5}
& \textbf{59.8} & \textbf{46.3} & \textbf{34.6} \\
\bottomrule
\end{tabular}
\end{table}

\subsection{Offline Experiments}
In contrast to online evaluation, where models \textit{autonomously determine response timing}, our offline experiments on Ego4D, OmniStarPro-RNG, and SVBench operate under pre-defined decoding schedules. To ensure fair comparison, we strictly follow the evaluation protocols and metrics established in prior work~\cite{chen2024videollm,wu2024videollm,li2025lion}, despite their non-generative nature, where evaluation is limited to ground-truth verification. For OmniStarPro, we further go beyond conventional perplexity and token-level accuracy checks by enabling full model generation with online-style scoring, which constitutes a key step toward comprehensive assessment of online capabilities.

\begin{table*}[t]
\centering
\caption{Evaluation results of various models on SVBench in dialogue and streaming evaluation. $\dag$ denotes fine-tuning on the SVBench training set.}
\label{tab:svbench}
\resizebox{0.9\textwidth}{!}{
\begin{tabular}{lccccccccccccc}
\toprule
\multirow{2}{*}{\textbf{Method}} & \multicolumn{6}{c}{\textbf{Dialogue Evaluation}} & \multicolumn{6}{c}{\textbf{Streaming Evaluation}} & \multirow{2}{*}{\textbf{AVG}} \\
\cmidrule(lr){2-7} \cmidrule(lr){8-13}
 & \textbf{SA} & \textbf{CC} & \textbf{LC} & \textbf{TU} & \textbf{IC} & \textbf{OS} & \textbf{SA} & \textbf{CC} & \textbf{LC} & \textbf{TU} & \textbf{IC} & \textbf{OS} & \\
\midrule
\multicolumn{14}{c}{\textit{Closed-source LVLMs}} \\
\midrule
GPT-4V & 56.03 & 62.61 & 69.09 & 65.36 & 53.73 & 60.30 & 56.37 & 61.41 & 65.80 & 59.18 & 57.16 & 57.93 & 59.12 \\
GPT-4o & 58.26 & 64.76 & 70.75 & 67.68 & 55.82 & 62.57 & 57.99 & 63.52 & 67.72 & 60.18 & 59.25 & 59.97 & 61.27 \\
\midrule
\multicolumn{14}{c}{\textit{Open-source Video-LLMs/LVLMs}} \\
\midrule
LLaVA-NeXT-Video & 37.71 & 44.59 & 52.05 & 41.80 & 36.58 & 41.40 & 34.29 & 39.68 & 47.65 & 35.33 & 36.68 & 36.12 & 38.76 \\
InternVL2.5 & 43.73 & 50.70 & 56.61 & 55.03 & 43.46 & 48.73 & 40.44 & 48.34 & 52.84 & 46.93 & 45.10 & 45.04 & 46.89 \\
InternVideo2.5 & 46.83 & 53.48 & 58.22 & 58.91 & 47.02 & 51.73 & 41.76 & 49.72 & 53.25 & 48.44 & 47.10 & 46.58 & 49.16 \\
MiniCPM-V 2.6 & 51.70 & 59.50 & 65.33 & 61.72 & 50.09 & 56.63 & 46.44 & 52.73 & 58.35 & 53.48 & 48.32 & 49.67 & 53.15 \\
Qwen2.5-VL & 52.54 & 59.85 & 65.52 & 64.64 & 51.23 & 57.57 & 48.21 & 56.12 & 60.31 & 56.33 & 52.46 & 52.84 & 55.21 \\
\midrule
\multicolumn{14}{c}{\textit{Online Assistants}} \\
\midrule
\rowcolor{gray!15} LiveStar & 46.43 & 53.75 & 59.36 & 57.29 & 45.64 & 51.37 & 43.56 & 51.52 & 55.71 & 50.79 & 47.77 & 48.15 & 49.76 \\
\rowcolor{gray!15} LiveStarPro & 47.49 & 56.58 & 63.86 & 59.70 & 48.92 & 53.72 & 44.33 & 55.49 & 59.83 & 52.87 & 50.48 & 50.68 & 52.20 \\
\rowcolor{gray!15} LiveStar$^\dag$ & 54.06 & 61.08 & 66.43 & 66.06 & 52.67 & 58.95 & 52.19 & 59.00 & 62.85 & 58.35 & 54.95 & 55.87 & 57.41 \\
\bottomrule
\end{tabular}
}
\end{table*}

\textbf{Evaluation Metrics.}
For standardized offline benchmarking, we adopt widely used metrics from prior studies, including PPL, TokAcc (Token Accuracy, previously termed LM-Correctness), TimeDiff, and Fluency~\cite{chen2024videollm,wu2024videollm,li2025lion}. In addition, we evaluate model outputs generated under offline-prescribed timing using online-style generation metrics, namely SemCor and SumFluen, both of which are scored with GPT-4o.

\begin{table}[t]
  \centering
  \caption{Offline evaluation on the Ego4D Narration Stream benchmark. ``-'' denotes incompatibility with the Fluency metric due to the absence of EOS tokens.}
  \label{tab:ego4d}
  \renewcommand{\arraystretch}{1}
  \setlength{\tabcolsep}{3pt}

  \begin{tabular}{lcccc}
    \toprule
    \textbf{Method} & \textbf{PPL}$\downarrow$ & \textbf{TimeDiff}$\downarrow$ & \textbf{Fluency}$\uparrow$ & \textbf{TokAcc}$\uparrow$ \\
    \midrule
    VideoLLM-online & 2.43 & 2.04 & 45.1\% & 48.1\% \\
    VideoLLM-MoD    & 2.41 & 2.04 & 45.2\% & 48.9\% \\
    LION-FS         & 2.09 & 2.15 & \textbf{46.1\%} & 52.4\% \\
    MMDuet          & 4.51 & 1.97 & - & 39.3\% \\
    \rowcolor{gray!15}
    \textbf{LiveStar} & 1.97 & 1.76 & - & 61.1\% \\
    \rowcolor{gray!15}
    \textbf{LiveStarPro} & \textbf{1.95} & \textbf{1.63} & - & \textbf{61.9\%} \\
    \bottomrule
  \end{tabular}
\end{table}

\textbf{Results.}
The left side of Tab.~\ref{tab:rng_evaluation} reports offline evaluation results on OmniStarPro-RNG under fixed decoding timing conditions. The results indicate that LiveStarPro outperforms all other online assistants as well as open-source offline Video-LLMs/LVLMs across all evaluated metrics, although a noticeable gap remains relative to human performance and GPT-4V/4o.
Notably, the offline portion of Tab.~\ref{tab:rng_evaluation} also offers a partial controlled signal for the benefit of the SVeD+SCAM paradigm: the InternVideo2.5 backbone evaluated without streaming fine-tuning achieves a SemCor of 4.32, whereas LiveStarPro---built on the same backbone---attains 4.62 under identical offline decoding conditions, a gain that cannot be attributed to backbone differences and therefore reflects the improved video-language alignment afforded by the SCAM training objective.
Moving to dialogue and streaming capabilities, Tab.~\ref{tab:svbench} presents the evaluation on SVBench. In the zero-shot setting, LiveStarPro attains an average of 52.20, which exceeds both its InternVideo2.5 backbone (49.16) and the streaming counterpart LiveStar (49.76), and approaches the strongest general-purpose open-source LVLMs such as Qwen2.5-VL (55.21) and MiniCPM-V 2.6 (53.15); the remaining gap is expected, since these baselines are trained on substantially larger instruction-tuning corpora than the streaming-oriented data used here. Once fine-tuned on the SVBench training set, LiveStar$^\dag$ improves its zero-shot average from 49.76 to 57.41 (a 15.37\% relative gain), thereby surpassing all open-source Video-LLMs and approaching the closed-source GPT-4V, which indicates that the streaming-oriented design does not preclude competitive interactive QA when in-domain supervision is available.
Finally, as shown in Tab.~\ref{tab:ego4d}, LiveStarPro consistently exceeds the performance of other online assistants on the Ego4D benchmark, attaining an 18.1\% higher TokAcc relative to the second-best LION-FS.

\subsection{Conventional Offline Experiments}
Although LiveStarPro is optimized for proactive online streaming, we must verify that its specialized design does not compromise general video understanding. Conventional benchmarks such as MVBench~\cite{li2024mvbench}, LongVideoBench~\cite{wu2024longvideobench}, and VideoMME~\cite{fu2025video} operate under an \textit{offline paradigm} in which models access the complete video file without real-time latency constraints, revealing whether the model retains spatial-temporal reasoning and long-context comprehension after streaming fine-tuning.

\textbf{Results.}
Table~\ref{tab:traditional_benchmarks} reports the performance of LiveStarPro on three representative offline benchmarks, alongside state-of-the-art offline LVLMs and online-capable models. LiveStarPro attains 69.8\% on MVBench, 56.3\% on LongVideoBench, and 60.8\% on VideoMME (w/o subtitles); relative to the InternVideo2.5 backbone it incurs a modest decline rather than a collapse, indicating that the proposed memory and training strategy largely preserve, rather than fully retain, offline reasoning ability.
Compared with representative offline models such as LLaVA-OneVision, LiveStarPro reaches higher accuracy on MVBench and stays close to recent large-scale vision-language models such as Qwen2.5-VL and InternVL2.5, while still trailing the strongest of them on long-context benchmarks. Among methods that support online inference, LiveStarPro consistently outperforms prior approaches and exceeds VideoChat-Online by 8.0 percentage points on VideoMME. Taken together, these results suggest that the hierarchical memory and the streaming-oriented training strategy keep the offline degradation commonly observed in online video assistants within an acceptable margin, while delivering the proactive streaming capability that the offline models lack.

\begin{table}[t]
\centering
\caption{Performance comparison on conventional offline video understanding benchmarks.}
\label{tab:traditional_benchmarks}
\renewcommand{\arraystretch}{1.0}
\setlength{\tabcolsep}{6pt}
\begin{tabular}{l c c c}
\toprule
\textbf{Model} & \textbf{MVBench} & \textbf{LongVideoBench} & \textbf{\makecell{VideoMME \\ (w/o sub.)}} \\
\midrule
\multicolumn{4}{c}{\textit{Offline Video-LLMs / LVLMs}} \\
\midrule
GPT-4V & 43.7 & 59.1 & 59.9 \\
GPT-4o & 64.6 & \textbf{66.7} & \textbf{71.9} \\
Gemini-1.5-Pro & 60.5 & 64.0 & 71.9 \\
LLaVA-NeXT-Video & 53.1 & 49.1 & 46.5 \\
LLaVA-OneVision & 56.7 & 56.3 & 58.2 \\
VideoLLaMA2 & 54.6 & - & 47.9 \\
Qwen2.5-VL & 69.6 & 56.0 & 65.1 \\
InternVL2.5 & \textbf{72.0} & 60.0 & 64.2 \\
\midrule
\multicolumn{4}{c}{\textit{Online Assistants}} \\
\midrule
MovieChat & 55.1 & 56.3 & 38.2 \\
VideoChat-Online & 64.9 & - & 52.8 \\
Dispider & - & - & 57.2 \\
\rowcolor{gray!15}
\textbf{LiveStarPro} & \textbf{69.8} & \textbf{56.3} & \textbf{60.8} \\
\bottomrule
\end{tabular}
\end{table}

\subsection{Ablation Study}
\begin{figure}[t]
\centering
\includegraphics[width=0.9\columnwidth]{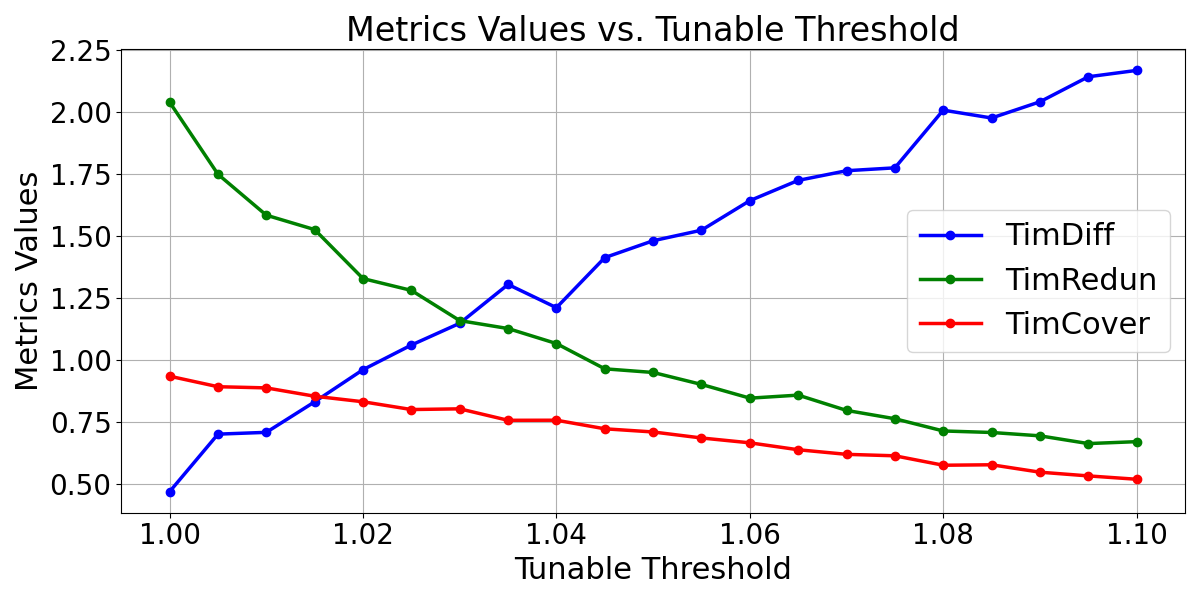}
\vspace{-2mm}
\caption{\textbf{Ablation study on the impact of response-silence threshold.}}
\label{fig:abl_alpha}
\end{figure}

\textbf{Impact of Response-Silence Threshold.}
LiveStarPro adopts a dynamic response-silence decoding scheme that is governed by an adaptive threshold. Specifically, the decoding threshold is formulated as \( \alpha \cdot \text{PPL}^{t_i}([Dec]) \), where \( \alpha \geq 1 \) denotes a tunable scaling factor applied to the timestep-dependent perplexity. To systematically examine the sensitivity of the model to \( \alpha \), we perform an empirical study over the range \( [1.0, 1.1] \) on the OmniStarPro-RNG benchmark. Results from online evaluation (Fig.~\ref{fig:abl_alpha}) highlight the pivotal role of \( \alpha \) in balancing timing difference, timing redundancy, and timing coverage. The experiments indicate that optimal performance is achieved within a narrow interval of \( \alpha = 1.02 \)–\( 1.04 \), and we select \( \alpha = 1.03 \) as the default setting.
The narrowness of this optimal range reflects a well-understood property of perplexity-based thresholds: because perplexity is computed relative to the LM's own probability distribution, its absolute scale varies across model families and domains, and $\alpha$ essentially normalizes a relative drift rather than an absolute magnitude. In deployment, $\alpha$ can therefore be calibrated efficiently on a small held-out stream (typically fewer than 100 frames) to identify the optimal operating point for a given domain, and this calibration cost is negligible compared to the main training procedure.

\textbf{Effect of Memory and Caching Strategies.}
Tab.~\ref{abl:strategies} summarizes ablation results for different memory management schemes and KV cache strategies. For memory strategies: (1) Uniform dropout leads to a 4.70\% decrease in SemCor due to the indiscriminate removal of critical recent frames; (2) FIFO-based forgetting disrupts temporal reasoning by discarding historical event captions, resulting in a 9.42\% increase in TimDiff and a 3.76\% reduction in SemCor; (3) Our Peak-End Memory Compression retains semantic clip summaries and keyframes via precomputed PPL-based dropout, achieving the best SemCor and the lowest TimDiff among short-term strategies; (4) Furthermore, incorporating TSHM for long-term retrieval boosts SemCor to 3.27, demonstrating that preserving evicted history in a recursive tree effectively mitigates catastrophic forgetting with only a marginal FPS trade-off.
For KV caching, disabling both inter-/intra-dialogue caching or intra-dialogue caching alone yields suboptimal throughput. In contrast, for the short-term Peak-End configuration, enabling both strategies improves inference throughput by 1.53$\times$ over no caching (3.82 vs. 2.50 FPS) and by 1.31$\times$ over intra-dialogue caching alone (3.82 vs. 2.92 FPS) under 5-minute video inference; the full LiveStarPro with TSHM further reaches 3.96 FPS, a 1.58$\times$ speedup over no caching. This avoids redundant recomputation of historical representations, thereby supporting low-latency inference without compromising output quality.

\begin{table}[t]
\footnotesize
\centering
\caption{Ablation study on memory strategies and KV cache for inference on OmniStarPro-RNG.}
\label{abl:strategies}

\setlength{\tabcolsep}{3.5pt}
\renewcommand{\arraystretch}{0.95}

\begin{tabular}{@{}c|c|ccc@{}}
\toprule
\textbf{Memory Strategy}  & \textbf{KV Cache}     & \textbf{SemCor$\uparrow$} & \textbf{TimDiff$\downarrow$} & \textbf{FPS$\uparrow$} \\
\midrule
Uniform Dropout           & \multirow{2}{*}{Both} & 3.04            & 2.01              & 3.77         \\
FIFO Forgetting           &                       & 3.07            & 2.09             & 3.91         \\
\midrule
\multirow{3}{*}{Peak-End (Short-term)}
                          & Neither               & 3.19            & 1.95             & 2.50         \\
                          & w/o Inter-Dialog      & 3.17            & 1.87             & 2.92         \\
                          & Both                  & 3.19            & 1.91             & 3.82         \\
\midrule
TSHM (Long-term) & Both                & 3.27   & 1.89    & 3.96         \\
\bottomrule
\end{tabular}
\end{table}

\textbf{Long-term Retrieval Diagnostics.}
We further dissect the long-term retrieval memory of TSHM on the LMR task of the OmniStarPro-Long partition, where the supporting evidence of a query frequently resides beyond the active context window. Tab.~\ref{tab:abl:retrieval} contrasts the recursive event tree against a flat $k$-nearest-neighbor memory bank that stores every evicted unit without hierarchical organization; the recall accuracy of the recursive event tree therefore coincides with the LMR row of Tab.~\ref{tab:OmniStarLong}, while the two baselines isolate the contribution of the hierarchical organization. The flat index attains a comparable recall on the short memory-span bucket, yet it degrades markedly on the long bucket and incurs a retrieval latency that grows linearly with the number of stored units. By contrast, the recursive event tree sustains a substantially higher recall on the long bucket and a sublinear retrieval latency, which empirically corroborates the logarithmic-retrieval property established in Sec.~\ref{sec:tshm_theory}. The benefit originates from the parent and child traversal, which recovers coherent event chains rather than isolated units.

\begin{table}[t]
\footnotesize
\centering
\caption{Long-term retrieval diagnostics on LMR of OmniStarPro- Long. Recall accuracy (\%) on the short/long memory-span buckets and average per-query retrieval latency (ms).}
\label{tab:abl:retrieval}
\setlength{\tabcolsep}{4pt}
\renewcommand{\arraystretch}{1}
\begin{tabular}{@{}lccc@{}}
\toprule
\textbf{Long-term Memory} & \textbf{Recall (S)$\uparrow$} & \textbf{Recall (L)$\uparrow$} & \textbf{Latency (ms)$\downarrow$} \\
\midrule
None (sliding window)     & 41.2 & 6.4  & --   \\
Flat $k$-NN bank          & 58.7 & 21.3 & 38.6 \\
\rowcolor{gray!15}
Recursive Event Tree      & \textbf{63.4} & \textbf{37.2} & \textbf{12.4} \\
\bottomrule
\end{tabular}
\end{table}

\textbf{Sensitivity of Tree Hyper-parameters.}
The structure of the recursive event tree is governed by the similarity threshold $\sigma$, which decides whether an evicted unit is attached as a child or initialized as a new root, and by the momentum factor $\beta$, which controls how rapidly a parent embedding evolves toward the centroid of its descendants. Tab.~\ref{tab:abl:tree} reports the long-bucket recall together with the resulting average branching factor and tree height as $\sigma$ and $\beta$ vary. A small $\sigma$ produces shallow and wide trees that collapse semantically distinct events into a single branch and thereby dilute retrieval precision, whereas an excessively large $\sigma$ fragments the memory into many isolated roots and reverts to a flat index. A moderate momentum factor allows parent embeddings to summarize their subtrees without erasing discriminative detail. The configuration $\sigma = 0.75$ and $\beta = 0.3$ yields the most favorable balance between branching factor and height and attains the highest long-bucket recall, and we adopt it as the default setting.

\begin{table}[t]
\footnotesize
\centering
\caption{Sensitivity of the recursive event tree to $\sigma$ and $\beta$. ``BF'' and ``H'' denote the average branching factor and tree height.}
\label{tab:abl:tree}
\setlength{\tabcolsep}{4pt}
\renewcommand{\arraystretch}{1}
\begin{tabular}{@{}cccc c@{}}
\toprule
$\sigma$ & $\beta$ & \textbf{BF} & \textbf{H} & \textbf{Recall (L)$\uparrow$} \\
\midrule
0.65 & 0.3 & 6.8 & 3.1 & 31.9 \\
0.75 & 0.1 & 3.9 & 5.4 & 34.7 \\
\rowcolor{gray!15}
0.75 & 0.3 & 4.1 & 5.2 & \textbf{37.2} \\
0.75 & 0.5 & 4.0 & 5.3 & 35.6 \\
0.85 & 0.3 & 2.2 & 8.7 & 33.1 \\
\bottomrule
\end{tabular}
\vspace{-1mm}
\end{table}






\subsection{Case Study}
\begin{figure*}[t]
\centering
\includegraphics[width=0.95\linewidth]{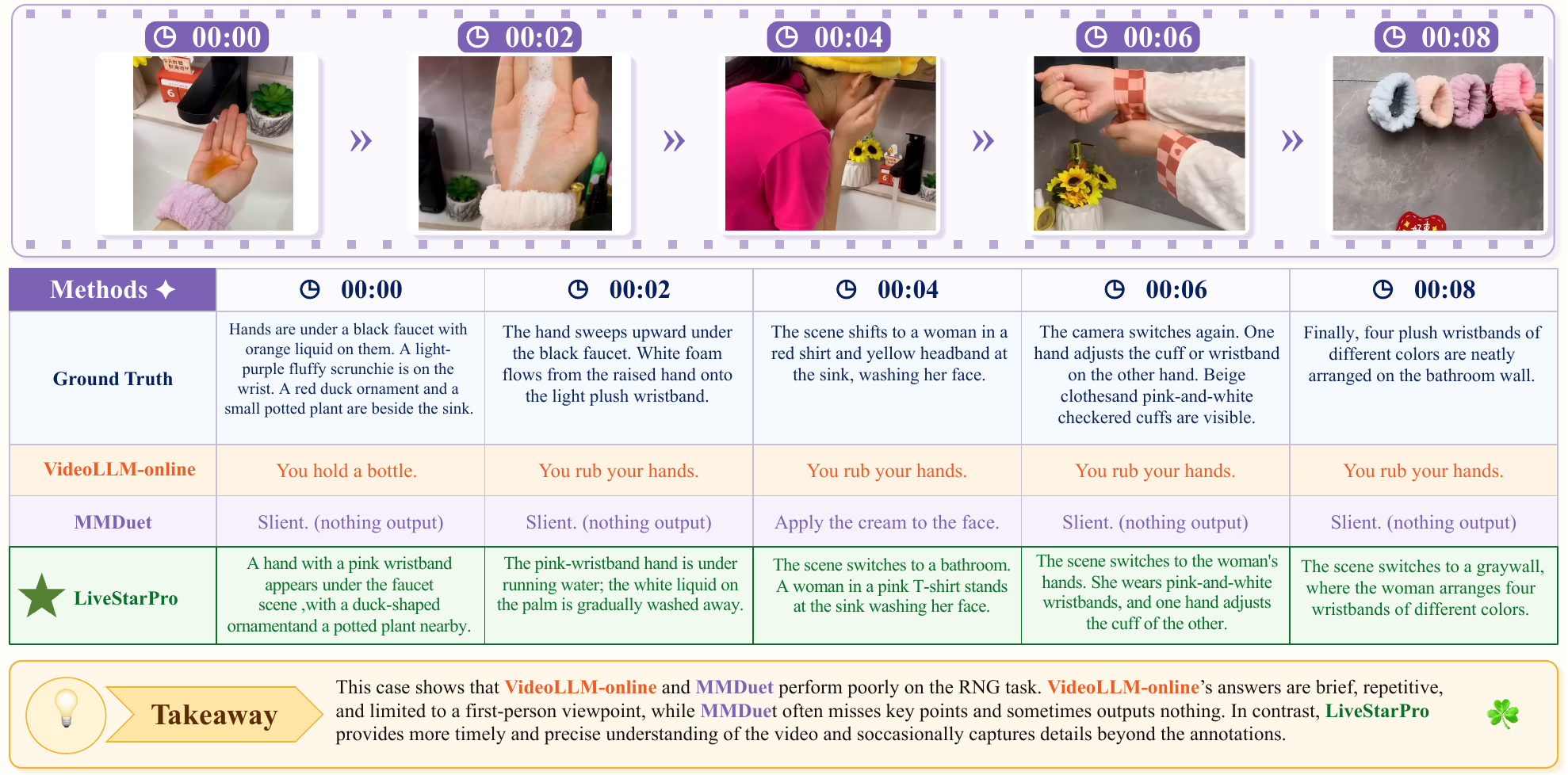}
\vspace{-1mm}
\caption{\textbf{Comparison on the RNG task.} LiveStarPro is timely and precise, while VideoLLM-online is repetitive and MMDuet often misses key points.}
\vspace{-2mm}
\label{fig:RNG_case}
\end{figure*}

We conduct qualitative comparisons on the RNG task between our LiveStarPro model and representative online video understanding baselines, namely VideoLLM-online and MMDuet. The qualitative results are illustrated in Fig.~\ref{fig:RNG_case}. These examples reveal that VideoLLM-online and MMDuet frequently exhibit limited contextual reasoning, hallucinated descriptions, and inadequate fine-grained recognition. By contrast, LiveStarPro consistently produces responses that are more accurate, more firmly grounded, and more temporally appropriate, owing to the effective integration of long-range context with streaming visual evidence. These qualitative comparisons underscore the advantage of LiveStarPro on fine-grained and context-sensitive online video understanding, while they also expose the limitations of existing LVLMs under realistic streaming settings.

\section{Conclusion}
This paper introduces LiveStarPro, a live streaming assistant that is conceived to deliver proactive responsiveness through adaptive streaming decoding. Around a streaming response--silence paradigm, we contribute three coordinated innovations: (1) a Streaming Verification Decoding (SVeD) mechanism that determines the appropriate response timing during real-time inference through single-pass perplexity verification, (2) a streaming video--language alignment framework that incorporates Streaming Causal Attention Masks (SCAM) during training, and (3) a Tree-Structured Hierarchical Memory (TSHM) that organizes evicted history into retrievable event chains for reasoning over long-horizon streams.
Extensive experiments across three benchmark datasets demonstrate that LiveStarPro attains state-of-the-art performance in online video understanding while preserving practical deployment efficiency, sustaining throughput of approximately 3 FPS on hour-long streams. On the OmniStarPro-Long partition, LiveStarPro further sustains reliable recall across all three memory-centric tasks (long-range memory recall, cross-event difference query, and temporal backtracking), confirming that TSHM effectively mitigates catastrophic forgetting for evidence that lies well beyond the active context window. By advancing a new response--silence paradigm together with the OmniStarPro benchmark, this work lays groundwork for robust, scalable models capable of long-horizon online video understanding.

\bibliography{LiveStarPro}
\bibliographystyle{IEEEtran}

\vfill

\end{document}